\documentclass[sigconf]{acmart}
\AtBeginDocument{%
  \providecommand\BibTeX{{%
    \normalfont B\kern-0.5em{\scshape i\kern-0.25em b}\kern-0.8em\TeX}}}

\copyrightyear{2023}
\acmYear{2023}
\setcopyright{acmlicensed}\acmConference[WWW '23]{Proceedings of the ACM Web Conference 2023}{May 1--5, 2023}{Austin, TX, USA}
\acmBooktitle{Proceedings of the ACM Web Conference 2023 (WWW '23), May 1--5, 2023, Austin, TX, USA}
\acmPrice{15.00}
\acmDOI{10.1145/3543507.3583509}
\acmISBN{978-1-4503-9416-1/23/04}

\usepackage{amsmath}
\usepackage{algorithm}
\usepackage{algorithmic}
\usepackage{amsfonts} 
\usepackage{xcolor}
\usepackage{subfigure}
\usepackage{multirow}
\usepackage{float}
\usepackage{multicol}
\usepackage{graphicx}
\usepackage{caption}

\usepackage{amsthm}
\newtheorem{definition}{Definition} 
\renewcommand{\shortauthors}{Zhang, et al.}
\begin{document}

\title{Minimum Topology Attacks for Graph Neural Networks}


% \IEEEauthorblockN{Mengmei Zhang}
% \IEEEauthorblockA{\textit{dept. Computer Science} \\
% \textit{Beijing University of Posts and Telecommunications}\\
% Beijing, China \\
% zhangmm@bupt.edu.cn}

\author{Mengmei Zhang}
\affiliation{%
  \institution{Beijing University of Posts and Telecommunications}
  \country{China}}
\email{zhangmm@bupt.edu.cn}

\author{Xiao Wang}
\affiliation{%
  \institution{Beijing University of Posts and Telecommunications}
  \country{China}}
\email{xiaowang@bupt.edu.cn}

\author{Chuan Shi}
\authornote{Corresponding author.}
\affiliation{%
  \institution{Beijing University of Posts and Telecommunications}
  \country{China}}
\email{shichuan@bupt.edu.cn}

\author{Lingjuan Lyu}
\affiliation{%
  \institution{Sony AI}
  \country{Japan}}
\email{lingjuan.lv@sony.com}

\author{Tianchi Yang}
\affiliation{%
  \institution{Beijing University of Posts and Telecommunications}
  \country{China}}
\email{yangtianchi@bupt.edu.cn}

\author{Junping Du}
\affiliation{%
  \institution{Beijing University of Posts and Telecommunications}
  \country{China}}
\email{junpingdu@126.com}

\begin{abstract}
  With the great popularity of Graph Neural Networks (GNNs), their robustness to adversarial topology attacks has received significant attention. Although many attack methods have been proposed, they mainly focus on fixed-budget attacks, aiming at finding the most adversarial perturbations within a fixed budget for target node. 
  However, considering the varied robustness of each node, there is an inevitable dilemma caused by the fixed budget, i.e., no successful perturbation is found when the budget is relatively small, while if it is too large, the yielding redundant perturbations will hurt the invisibility.
  To break this dilemma, we propose a new type of topology attack, named minimum-budget topology attack, aiming to adaptively find the minimum perturbation sufficient for a successful attack on each node. To this end, we propose an attack model, named MiBTack, based on a dynamic projected gradient descent algorithm, which can effectively solve the involving non-convex constraint optimization on discrete topology. Extensive results on three GNNs and four real-world datasets show that MiBTack can successfully lead all target nodes misclassified with the minimum perturbation edges. Moreover, the obtained minimum budget can be used to measure node robustness, so we can explore the relationships of robustness, topology, and uncertainty for nodes, which is beyond what the current fixed-budget topology attacks can offer.
%   Moreover, evaluating the adversarial robustness of each node amounts to finding the minimum perturbation for misclassification of node.
\end{abstract}




\maketitle

\section{Introduction}

Graph is commonly used to model many real-world relationships, such as social networks, citation networks, and e-commerce networks. 
Recently, there has been a surge of interest in Graph Neural Networks (GNNs) for representation learning of graphs, which combine node feature and topology with neural networks and have achieved outstanding performance in various application areas~\cite{Zhou2020GraphNN}. 

Despite the great success, GNNs have been proved to often suffer from adversarial attacks~\cite{nettack,adv_rl}, especially for topology attacks where the attacker tries to slightly manipulate topology to modify the decision of GNNs. Recent research has developed stronger topology attack methods~\cite{Xu2019TopologyAA,Zheng2021GraphRB,geisler2021robustness,Wang2022ClusterAQ,Wang2022BanditsFS} to explore the robustness of GNNs. Specifically, existing works all belong to the fixed-budget topology attacks, i.e., for each node, the attacker finds the most adversarial perturbations within a given budget (i.e., a fixed number of perturbed edges).
Usually, the budget of each node is heuristically specified as the same value or based on the node degree, and then it will keep unchanged once specified.

\begin{figure}[!t]
\centering
\includegraphics[width=1.0\linewidth]{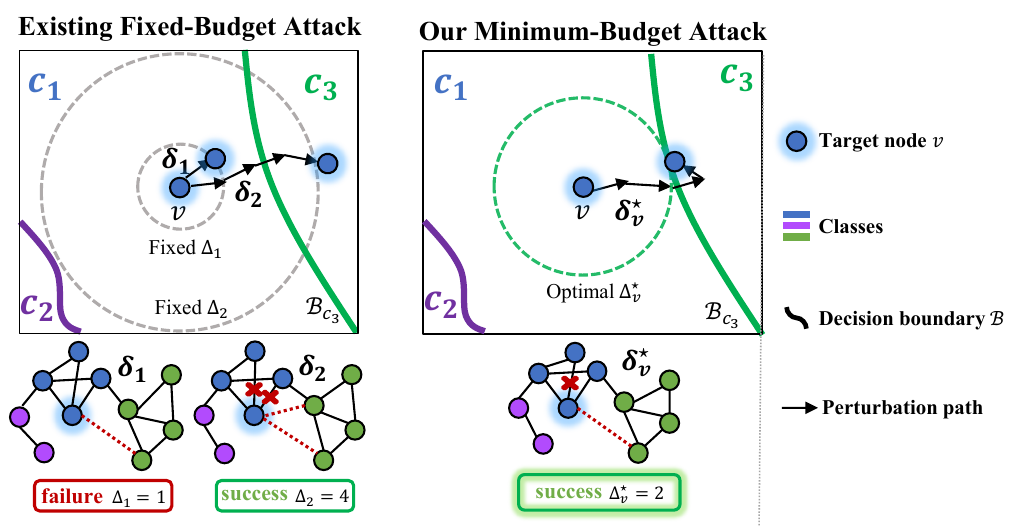}
\caption{{Illustrations of the existing fixed-budget topology attacks which may fail to cross decision boundary or cross too much, and our MiBTack can exactly cross the closest green decision boundary $\mathcal{B}_{c_3}$ with no waste. }}
\label{fig:butterfly}
\end{figure}

However, in the real-world graph, each node has unique feature and topology, and thus has varying adversarial robustness.
It naturally raises a fundamental question: \textit{Are the previous fixed-budget topology attacks truly suitable for evaluating the robustness of GNNs?} 
On the one hand, usually, an effective topology attack should lead to the misclassification of the target node and keeps it invisible, e.g., the topology cannot be modified too much. On the other hand, considering varied distribution of each node, evaluating the robustness of GNNs on node level can help better understand the robustness.
Unfortunately, the fixed-budget perturbation cannot satisfy the above requirements. First, for each node, the heuristically specified budget is usually not optimal: 
no successful perturbation can be found when the specified budget is too small, while for the large one, the redundant perturbations will hurt the invisibility.
As illustrated in Fig.~\ref{fig:butterfly}, given blue (class $c_1$) target node $v$, existing fixed-budget topology attacks aim to maximally change the prediction (color) of $v$ by modifying a fixed number of edges.  Only perturbing one edge (budget $\Delta_1=1$), the attacker fails to find a successful perturbation $\boldsymbol{\delta}_1$ to cross purple (class $c_2$) or green (class $c_3$) decision boundary. While perturbing four edges ($\Delta_2=4$) will cross the boundary too much, hurting the invisibility of the attack. 
The above analysis shows that there is a dilemma between a fixed budget and an effective attack.
Second, given the fixed budget, the total accuracy of all target nodes under attacks can be used to evaluate the robustness of GNNs, while evaluating the adversarial robustness of a sample (e.g., a node) amounts to finding the minimum perturbation for misclassification of it~\cite{Rony2019DecouplingDA}.

Clearly, the fixed budget restricts the potential of attack methods due to the different robustness of each node. Henceforth, we propose a new orthogonal minimum-budget topology attack, which aims at adaptively finding the minimum budget that is sufficient for a successful attack for each node. Thus the attacker can adapt to the robustness of each node as the green dotted line in Fig.~\ref{fig:butterfly}. 
Moreover, if we can adaptively find the budget for each node, we can provide a better understanding or more insights on node robustness, uncertainty, etc. 

In essence, the minimum-budget attacks are inherently different from the fixed-budget. The attacker with a fixed budget in Fig.~\ref{fig:butterfly} aims to flip $\Delta_2$ edges to maximally cross the decision boundary. Obviously, this is a combinatorial optimization problem constrained by $\Delta_2$. To solve it, exiting topology attacks often greedily flip edges for $\Delta_2$ times~\cite{nettack,adv_meta,adv_AAAI2020}. Further, some works use Projected Gradient Descent (PGD) to project the combination of perturbed edges within the $\Delta_2$-limited space (grey dotted circle) ~\cite{Sun2018DataPA,Xu2019TopologyAA,Zhang2021ProjectiveRA}. 
While the attacker with an adaptive budget will learn a  minimum perturbation (green dotted circle $\Delta_v^\star$) for crossing the decision boundary, which is challenging. With misclassification guarantee of non-convex GNN, the minimum-budget topology attack is essentially a non-convex constrained optimization problem, where PGD-based model can not be directly used~\cite{Pintor2021FastMA}. And the greedy-based model is too myopic to find a good solution~\cite{Liu2019AUF}. Grid search or binary search on budget for each node is also unbearable in practice.

In this paper,  we propose a novel \textbf{Mi}nimum-\textbf{B}udget \textbf{T}opology att\textbf{ack} method for GNNs named \textbf{MiBTack} based on dynamic projected gradient descent, in order to solve the above challenges.
Specifically, we decouple the budget and perturbation, turning the non-convex constrained optimization to alternatively solve the easier convex constrained optimization and update the budget separately.
In each iteration, we search for the most adversarial topology attacks under the current budget with PGD and then the budget is enlarged or reduced based on whether these attacks succeed. 
Through repeatedly crossing the green decision boundary, MiBTack can find the optimal budget $\Delta^\star_v$, namely the green dotted circle centered at node $v$ as shown in Fig.~\ref{fig:butterfly}, which is tangent to the green decision boundary.
Finally, our MiBTack can successfully attack (0 ACC) all target nodes with minimum budget.

The contributions of this paper are summarized as follows:

$\bullet$ We highlight the inherent bottleneck of existing fixed-budget topology attacks, then we propose an orthogonal type of topology attack for GNN, which aims to adaptively find minimum budget for a successful attack. The proposed attacks are highly threatening and can be used to quantify the node-level adversarial robustness.

$\bullet$ We propose an effective minimum-budget attack model MiBTack based on a differentiable dynamic projected gradient descent module, in order to solve the involving non-convex constraint on discrete topology. 

$\bullet$ Extensive experiments on four real-world datasets and three GNNs show that our MiBTack can produce successful attacks with minimum perturbed edges.  With the obtained minimum budgets, we also explore the relationships between robustness, topology and uncertainty.

\section{Related Work}\label{app:related}
% GNN
\textbf{Graph Neural Networks.}
Graph analysis has attracted considerable attention as graphs exist in various complex systems~\cite{Wang2017CommunityPN, Sun2023BHGNNRTNE}.
GNNs \cite{GCN,Zhou2020GraphNN,Liu2022MGNNIMG,Liu2022EIGNNEI} on graph structured data have shown outstanding results in various tasks. 
A special form of GNNs is graph convolutional networks (GCNs) \cite{GCN}, which learn on graph structures using convolution operations and achieve state-of-the-art performance. To further improve GCNs, \cite{Klicpera2019PredictTP} derives an improved propagation scheme based on personalized PageRank, which further leverages a large, adjustable neighborhood for classification, and \cite{Wu2019SimplifyingGC} further simplifies GCN by removing the non-linearities between GCN layers, etc. In this paper, we mainly focus on attacking GCN and its variants.

\textbf{Topology Attacks.}
With the wide applications of GNNs, their robustness to adversarial attacks has received increasing attention, especially for topology attacks  \cite{nettack,adv_rl,Xu2019TopologyAA,adv_AAAI2020,geisler2021robustness,Wang2022ClusterAQ,Wang2022BanditsFS,Zhang2022UnsupervisedGP,adv_rewring,Zhang2022UnsupervisedGP,Liu2022SurrogateRL,Yang2021DerivativefreeOA,Hussain2021StructackSA}. They try to find the most adversarial perturbations within a fixed budget $\Delta$. To avoid the combinatorial search in discrete space, a basic solution is to greedily select perturbed edges with top-$\Delta$ adversarial scores~\cite{nettack,Wu2019AdversarialEF}. The advanced works suggest utilizing gradient to search for optimal combinations. Specifically, they often use projected gradient descent (PGD) to project the updated perturbations into the $\Delta$-constrained space \cite{Sun2018DataPA,Xu2019TopologyAA,geisler2021robustness}. However, they mainly focus on fixed-budget topology attacks.
We introduce the first study on a new minimum-budget topology attack. In addition, different from the previous simple convex constraint on budget, we need to solve an  intractable non-convex constraint on GNN which cannot be directly solved by PGD~\cite{Pintor2021FastMA}.

% minimum budget attacks. 
\textbf{Minimum-Norm Attacks.}
Adversarial attacks on machine learning can be categorized into maximum-confidence attacks and minimum-norm attacks~\cite{Pintor2021FastMA}. Given the target image, the former tries to find the perturbation $\boldsymbol{\delta}$ within a given bound of the $\boldsymbol{\delta}$'s $L_p$ norm to maximize the confidence of misclassification, e.g, FGSM~\cite{Goodfellow2015ExplainingAH}. Recently, the latter is proposed to find the perturbation with minimum $L_p$ norm sufficient for misclassification~\cite{Rony2019DecouplingDA,Carlini2017TowardsET,Pintor2021FastMA}. In contrast to the former attacks that can only be used to evaluate model robustness, the minimum-norm attacks can also be used to measure an image's robustness, i.e., the distance of the image to the decision boundary. Intuitively, a more robust image requires a larger perturbation to cross the decision boundary~\cite{Qin2020ImprovingCT}. Thus the minimum-norm attacks are often used to better understand the adversarial robustness and its inherent relations to underlying data distribution~\cite{Qin2020ImprovingCT,Carlini2019DistributionDT}. However, there are no existing minimum-norm attacks for discrete graph data, and this paper sheds the first light on this important problem.

\section{Preliminaries}
In this section, we briefly introduce the backgrounds and preliminaries of GNNs and existing topology attacks. We summarize the related works in Appendix~\ref{app:related}.

\subsection{Graph Neural Networks}
% Recently, it has been shown that GNNs are powerful in a range of tasks~\cite{Zhou2020GraphNN}. 
We consider attacking GNNs on the semi-supervised node classification task. Formally, we define an undirected graph as $G = (\mathcal{V}, \mathcal{E}, \boldsymbol{X})$, where $\mathcal{V}$ is the set of $N$ nodes, $\mathcal{E}$ represents the set of edges, and $\boldsymbol{X}= [\boldsymbol{x}_{1} ; \boldsymbol{x}_{2} ; \cdots ; \boldsymbol{x}_{N}] \in \mathbb{R}^{N \times D}$ is the node feature matrix. We denote $\boldsymbol{a}_{v}=\left[a_{v 1} ; a_{v 2} ; \cdots ; a_{v N}\right] \in \{0,1\}^{N}$ as the adjacency vector of node $v$.
Let $d_v$ and $\mathcal{N}(v)$ represent the degree and the neighborhood set of $v$, and the $\tilde{d}_v$ and $\tilde{\mathcal{N}}(v)$  are the ones including self-loop $(v,v)$. In this paper, we focus on node classification, where each node $v \in \mathcal{V}_L$ in training set $\mathcal{V}_L \subset \mathcal{V}$ has a label $y_{v} \in \mathcal{Y}$ and $\mathcal{Y}=\{1,2, \cdots, C\}$. Given $\mathcal{V}_L$, the goal of node classification task is to predict the class of node in unlabeled data $\mathcal{V}_U$. Taking the representative GCN~\cite{GCN} as an example, GCN uses convolution operations to aggregate neighboring nodes as follows:
\begin{equation}
    \boldsymbol{h}_{v}^{(l)}=\operatorname{ReLU}(\boldsymbol{W}^{(l)}(\sum_{u \in \tilde{\mathcal{N}}(v)} \tilde{d}_{u}^{-1 / 2} \tilde{d}_{v}^{-1 / 2} \boldsymbol{h}_{u}^{(l-1)})),
\end{equation}
where $\boldsymbol{h}^{(l)}_v$ is the representation of node $v$ at the $l-$th layer, and  $\boldsymbol{h}^{(0)}_v=\boldsymbol{x}_v$ for initialization. $\boldsymbol{W}^{(l)}$ is the trainable weight matrix in the $l-$th layer, and we set $\boldsymbol{\theta} = \{\boldsymbol{W}^{(1)}, \cdots, \boldsymbol{W}^{(L)}\}$ as all model parameters. The output of node $v$ is  $f_{\boldsymbol{\theta}}(\boldsymbol{a}_{v})=\operatorname{softmax}(\boldsymbol{h}_{v}^{(L)}) \in [0,1]^{C}$, where $f_{\boldsymbol{\theta}}(\boldsymbol{a}_{v})_{c}$ %\textcolor{red}{what is $a_v$} 
indicates the probability of node $v$ being classified into class $c$, i.e., confidence. Then, the model parameters $\boldsymbol{\theta}$ are learned by minimizing the cross-entropy loss on the outputs of the training nodes $\mathcal{V}_{L}$ as follows:
\begin{equation}\label{eq:gnn_training}
    \boldsymbol{\theta}^{*}=\arg \min _{\boldsymbol{\theta}}-\sum_{v \in \mathcal{V}_{L}} \ln f_{\boldsymbol{\theta}}(\boldsymbol{a}_{v})_{y_v}.
\end{equation}
%Following \cite{GCN}, we consider GCN with two layers ($L=2$) in this paper.

\subsection{Adversarial Topology Attacks}

Here we focus on adversarial topology attacks under the following settings:

$\bullet$ {Attack Goal.} Our attacker's goal is to lead GNN misclassify the target node $v$ by manipulating its adjacency vector $\boldsymbol{a}_{v}$.

$\bullet$ Attack Type. We mainly focus on evasion attacks, where the parameters of the trained model are assumed to be fixed when being attacked. We also perform experiments for the challenging poisoning case,  where the model is retrained after the attack.
% Once the model is deployed, the attacker can launch an unnoticeable evasion attack at any time, which increases the difficulty and cost of detection.

$\bullet$ {Attack Knowledge.}
We consider the worst case, in which an attacker can access the internal configurations (i.e., the learned parameters) of the targeted GNN model. This assumption ensures reliable vulnerability analysis in the worst case.

Formally,  given a target node $v$, the attackers under these settings aim to decrease the performance of the well-trained GNN $f_{\boldsymbol{\theta}^*}$ on node $v$. They optimize a topology perturbation vector $\boldsymbol{\delta}_v \in [\delta_{v1}, \delta_{v2}, \cdots, \delta_{vN}] \in \{0,1\}^N$ for perturbing adjacency vector $\boldsymbol{a}_v$, where the $u$-th element $\delta_{vu}=1$ means flipping the status of edge $(v,u)$ (adding or deleting). Specifically, we add perturbation to $\boldsymbol{a}_v$ by  $\boldsymbol{a}^{\prime}_v = \boldsymbol{a}_v + \mathcal{T}(\boldsymbol{\delta}_v) = \boldsymbol{a}_v + (\boldsymbol{1}-2\boldsymbol{a}_v) \odot \boldsymbol{\delta}_v$, where $\odot$ denotes element-wise multiplication. So element $\delta_{vu}=1$ indicates adding edge $(v,u)$ when ${a}_{vu}=0$, deleting edge $(v,u)$ when ${a}_{vu}=1$. 

\textbf{Fixed-budget Topology Attacks.}  Existing topology attacks mainly focus on fixed-budget topology attacks. Given a target node $v$, they often specify the maximum amount of perturbation edges in advance, then try to find a  perturbation vector $\boldsymbol{\delta}_v$ to maximally degrade GNN's performance on $v$ with a constraint of $\|\boldsymbol{\delta}_v\|_0 \leqslant \Delta_v$ as follows:
\begin{align}\label{eq:fixed_budget}
    & \boldsymbol{\delta}^{\star}_v =\mathop{\rm arg min}\limits_{\boldsymbol{\delta}_v} \quad \mathcal{L}(\boldsymbol{a}^{\prime}_v) , \\
    & s.t.  \| \boldsymbol{\delta}_v\|_0 \leqslant \Delta_v, \quad \boldsymbol{\delta}_v \in \{0,1\}^N, \notag
\end{align}
where the attack loss $\mathcal{L}$ is the Carlili-Wagner (CW) loss following~\cite{Xu2019TopologyAA}:
\begin{equation}\label{eq:cw_loss}
     \mathcal{L}(\boldsymbol{a}^{\prime}_v) =   f_{\boldsymbol{\theta}^*}(\boldsymbol{a}^{\prime}_v)_{y_v} - \mathop{\rm max}\limits_{c \neq y_v}  f_{\boldsymbol{\theta}^*}(\boldsymbol{a}^{\prime}_v)_{c}, \\
\end{equation}
where the smaller $\mathcal{L}(\boldsymbol{a}^{\prime}_v)$ indicates the stronger attacks.  Clearly, as the constraints in Eq.~\eqref{eq:fixed_budget} showed, the fixed-budget attacks try to search the most adversarial $\Delta_v$ perturbation edges, which is inherently a combinatorial optimization problem. Existing works solve this problem mainly by two ways: (1) A straight solution is greedy search: the attacker greedily flips the edge which degrades loss $\mathcal{L}$ most and repeats it for $\Delta_v$ times.  
Doubtless, the greedy solver is myopic and will neglect some flipping actions that might be better in the long run~\cite{Liu2019AUF}. (2) So the recent works~\cite{Xu2019TopologyAA, Liu2019AUF} tend to relax $\boldsymbol{\delta}_v \in \{0,1\}^N$ to its convex hull $\boldsymbol{\delta}_v \in [0,1]^N$, yielding a continuous optimization problem. Then they can optimize $\boldsymbol{\delta}_v$ with gradient-based solvers which are proved to be much better than the classical greedy method. Specifically, they often use projected gradient descent (PGD) to solve the convex constraint $\boldsymbol{\delta}_v \in [0,1]^N$. 

As shown in the above objective, it is hard to generate both successful and unnoticeable attacks by the fixed budget attacks. Given a small budget, the perturbed target node may not be misclassified ($\mathcal{L}(\boldsymbol{a}^{\prime}_v)<0$),  while a large budget will hurt the invisibility of attacks. 

\section{Minimum-budget Topology Attack}

\subsection{Attack Objective}
Considering the mentioned inherent dilemma of existing fixed-budget topology attacks, we first propose an orthogonal minimum-budget topology attack for GNNs, which aims to adaptively find the minimum perturbation that is sufficient for the misclassification of target node. Formally, the objective of our attacks can be formulated as: 
\begin{equation}
   \begin{aligned}
     & \boldsymbol{\delta}^{\star}_v = \mathop{\rm argmin}\limits_{\boldsymbol{\delta}_v}  \quad \| \boldsymbol{\delta}_v \|_0, \\ \quad
    & s.t.\ \mathcal{L}(\boldsymbol{a}^{\prime}_v) <\gamma, \quad \boldsymbol{\delta}_v \in \{0,1\}^N, 
    \end{aligned}
    \label{eq:eq_our_obj_1}
\end{equation}
where $\mathcal{L}(\boldsymbol{a}^{\prime}_v)$ is CW attack loss defined in Eq.~\eqref{eq:cw_loss} and  $\boldsymbol{a}^{\prime}_v = \boldsymbol{a}_v + \mathcal{T}(\boldsymbol{\delta}_{v})$. $\gamma (\gamma >0)$ is the confidence level of misclassification, and a higher $\gamma$ will lead to crossing decision boundary more. In this paper, we set $\gamma=0$ by default, namely, generate the attacks which just cross the decision boundary exactly. So we can break the dilemma of fixed-budget about effectiveness and invisibility. Besides, $\boldsymbol{\delta}^{\star}_v$ can be used to evaluate the robustness of each node. Intuitively, if a node requires more perturbations to be successfully attacked, the node has a larger distance to the decision boundary and is more robust for topology attacks. Here we define the robustness ${\rho}_v$ of node $v$ as follows:
\begin{definition}
\textbf{Node Robustness.} Given a node $v$ and a graph neural network $f_{\boldsymbol{\theta}^*}$, the node robustness ${\rho}_v$ is defined as the amount of perturbed edges which are sufficient to make $f_{\boldsymbol{\theta}^*}$ misclassify node $v$ with minimum perturbation.
\end{definition}
Based on our attack objective, the node robustness can be calculated by ${\rho}_v = \|\boldsymbol{\delta}^{\star}_v\|_0$. Then we can use it to explore the relationships between robustness and data distribution, then provide a better understanding or more insights on node robustness.

Unfortunately, effectively and efficiently optimizing $\boldsymbol{\delta}^{\star}_v$ is challenging. Compared to the fixed-budget attack in Eq.~\eqref{eq:fixed_budget}, we can observe that our attack objective has an extra non-convex constraint $\mathcal{L}(\boldsymbol{a}^{\prime}_v) < \gamma$ (the non-convexity of GNN model $f_{\boldsymbol{\theta^*_v}}$). Solving such non-convex constrained optimization problem is harder. The advanced PGD-based model can only solve the convex constraint like $\|\boldsymbol{\delta}_v\|_0 \leqslant \Delta_v$, and is not readily applied in our attacks. Doubtlessly, the greedy-based model is too myopic to find a good solution.

\begin{figure*}[!t]
    \centering
    \includegraphics[width=0.75\linewidth]{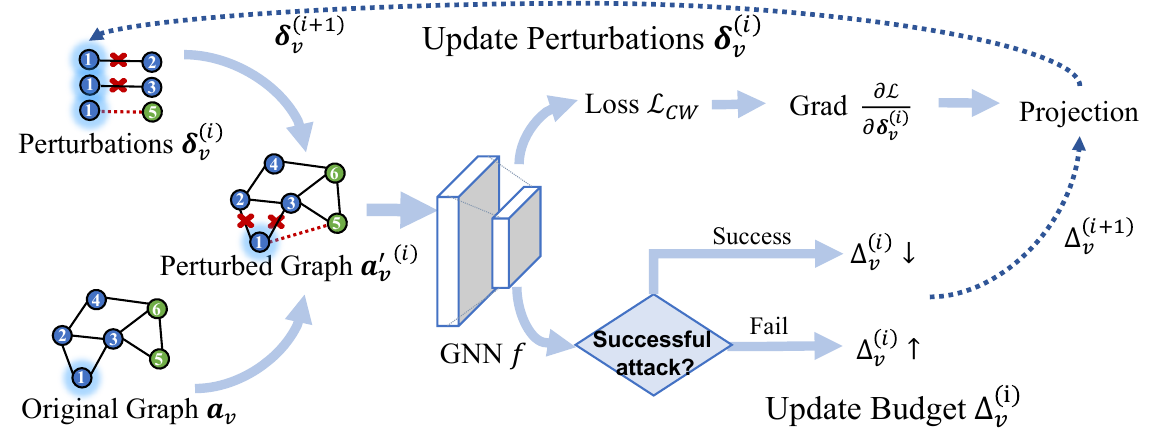}
    \vspace{-5pt}
    \caption{The overall framework procedure of our proposed MiBTack.
    In $i$-th iteration, MiBTack alternatively updates the current perturbation $\boldsymbol{{\delta}}_v^{(i)}$ and $\Delta_v^{(i)}$ with a dynamic PGD algorithm.
    }
    \label{fig:model}
\end{figure*}
\vspace{-5pt}

\subsection{The Proposed Model MiBTack}
Next, we describe the proposed minimum-budget attack model, MiBTack,  which solves the non-convex constrained optimization problem effectively. MiBTack is based on a dynamic PGD, which is mainly a projected gradient descent (PGD) method with a dynamic budget, so that the non-convex constrained optimization can be turned to alternatively solving the easier convex constrained optimization and updating budget separately.  Fig.~\ref{fig:model} shows the overall framework of MiBTack, which consists of two components:
(1) We fix current budget $\Delta_v$ and search for $\boldsymbol{{\delta}}_v$ under $\Delta_v$ which is a convex constrained optimization and can be solved by PGD. (2) If $\boldsymbol{{\delta}}_v$ can lead to a misclassification of target node $v$ (i.e., $\mathcal{L}(\boldsymbol{a}^{\prime}_v)<0$), the current $\Delta_v$ will be decreased for next iteration, otherwise increased. Thus the dynamic PGD can converge to $\mathcal{L}(\boldsymbol{a}^{\prime}_v)<0$ and lead to a finer search of a minimal budget by repeatedly crossing the decision boundary.

\textbf{Updating Perturbation.}
In this step, we aim to keep the budget fixed and update perturbation with the constraint of current budget. Specifically, for $i$-iteration, we fix $\Delta^{(i)}_v$ as a constant and update perturbations $\boldsymbol{{\delta}}^{(i-1)}_v$ by minimizing current attack loss $\mathcal{L}(\boldsymbol{a}^{\prime(i-1)}_v)$ within budget $\Delta^{(i)}_v$:
\begin{align}\label{eq:pgd_obj}
    & \min _{\boldsymbol{\delta}_{v}^{(i)}} \mathcal{L} \left(\boldsymbol{a}^{\prime(i-1)}_{v}\right), \\
    & \text { s.t. }\left\|\boldsymbol{\delta}_{v}^{(i)}\right\|_{0} \leqslant \Delta_{v}^{(i)}, \boldsymbol{\delta}^{(i)}_v \in [0,1]^N, \notag
\end{align}
where $\boldsymbol{a}^{\prime(i-1)}_{v}=\boldsymbol{a}_{v}+\mathcal{T}(\boldsymbol{{\delta}}_v^{(i-1)})$. Following~\cite{Xu2019TopologyAA}, we also relax the discrete $\boldsymbol{\delta}^{(i)}_v \in \{0,1\}^N$ to continue $[0,1]^N$ for ease of gradients. As we can see, this is the easier convex constrained problem which can be solved through PGD, which contains gradient descent and projection. The idea of PGD is as follows: if the perturbation variable after the gradient descent update is out of current budget $\Delta^{(i)}_v$, PGD will project it back to the $\Delta^{(i)}_v$-constrained space.  In gradient descent, we calculate the gradient $\boldsymbol{g}$ of the attack loss w.r.t. $\boldsymbol{{\delta}}_v^{(i-1)}$:
\begin{equation}\label{eq:PGD_grad}
    \mathbf{g} \leftarrow \nabla_{\boldsymbol{{\delta}}_v^{(i-1)}} \mathcal{L}(\boldsymbol{a}^{\prime(i-1)}_{v}).
\end{equation}
Given $\boldsymbol{g}$, we update $\boldsymbol{\delta}_v^{(i-1)}$ with the normalized gradient descent:
\begin{equation}\label{eq:PGD_gd}
    \Tilde{\boldsymbol{\delta}}^{(i)}_v \leftarrow \boldsymbol{\delta}^{(i-1)}_v+\alpha \cdot \boldsymbol{g} /\|\boldsymbol{g}\|_{2},
\end{equation}
where $\alpha$ is the step size of gradient descent. Once $\boldsymbol{\delta}^{(i-1)}_v$ is updated to $\Tilde{\boldsymbol{\delta}}^{(i)}_v$, to keep $\|{\boldsymbol{\delta}}^{(i)}_v\|_0 \leqslant \Delta^{(i)}_v$ and $\boldsymbol{\delta}^{(i)}_v \in [0,1]^N$, we project $\Tilde{\boldsymbol{\delta}}^{(i)}_v$ via a projection operator:
\begin{equation}\label{eq:PGD_proj}
    \boldsymbol{\delta}_{v}^{(i)} \leftarrow \operatorname{Proj}_{[0,1]^{N}}[\Tilde{\boldsymbol{\delta}}_{v}^{(i)}-\mu \mathbf{1}],
\end{equation}
where the perturbation vector $\Tilde{\boldsymbol{\delta}}^{(i)}_v$ encodes the score of flipping edges of $v$. For $u$-th element, a larger value of  $\Tilde{\delta}^{(i)}_{vu}$ indicates a stronger attack effect of flipping edge $(v,u)$. To fulfill the constraint of $\Delta^{(i)}_v$, we denote the ($\Delta^{(i)}_v$+1)-th largest value of $\Tilde{\boldsymbol{\delta}}^{(i)}_v$ as a scalar $\mu$, so only the most $\Delta^{(i)}_v$ perturbation edges are kept non-negative values in $\Tilde{\boldsymbol{\delta}}_{v}^{(i)}-\mu \mathbf{1}$, otherwise negative. Then we use clip operation ${\rm Proj}_{[0,1]^N}$ to set these negative elements as zero. Thus, after projection, $\boldsymbol{\delta}_{v}^{(i)}$ can fulfill the constraints $\left\|\boldsymbol{\delta}_{v}^{(i)}\right\|_{0} \leqslant \Delta_{v}^{(i)}$ and $\boldsymbol{\delta}^{(i)}_v \in [0,1]^N$ in Eq.~\eqref{eq:pgd_obj}, and can be used to perturb the topology of $v$ by  $\boldsymbol{a}_{v}^{\prime(i)} \leftarrow \boldsymbol{a}_{v}+\mathcal{T}(\boldsymbol{\delta}_{v}^{(i)})$.

\textbf{Updating Budget.}
In this step, based on the above updated perturbation  $\boldsymbol{\delta}^{(i)}_v$, we will dynamically adjust the current budget $\Delta^{(i)}_i$ depending on whether the current perturbation can successfully attack.

To identify whether the current perturbation can succeed, one direct way is to test the predicted label of the perturbed adjacency vector $\boldsymbol{a}^{\prime(i)}_v \in [0,1]^{N}$. While the real-world graph topology and its perturbations are discrete, so a more precise way is to  project $\boldsymbol{a}^{\prime(i)}_v$ from continuous space to discrete space ${\rm Proj}_{\{0,1\}^N}[\boldsymbol{a}^{\prime(i)}_v] \in \{0,1\}^{N}$ first. The projection method we use is straightforward: we choose the non-zero cells in perturbation $\boldsymbol{\delta}^{(i)}_v$ and turn them into 1. 
Next, if the  current perturbation $\boldsymbol{\delta}^{(i)}_v$ is sufficient to cross decision boundary (i.e., $\mathcal{L}({\rm Proj}_{\{0,1\}^N}[\boldsymbol{a'}^{(i)}_v])<0$), it can be determined that the current budget is large enough, and the optimal budget must be no more than $\Delta^{(i-1)}_v$, so we will decrease the budget $\Delta^{(i)}_v$ for next iteration with step size $\beta$ by:
\begin{equation}\label{eq:dyn_shrink}
    \Delta^{(i+1)}_v \leftarrow {\rm min}(\Delta^{(i)}_v-1,\Delta^{(i)}_v(1-\beta)),
\end{equation}
otherwise, we increase current $\Delta^{(i)}_v$ by:
\begin{equation}\label{eq:dyn_enlarge}
    \Delta^{(i+1)}_v \leftarrow {\rm max}(\Delta^{(i)}_v+1,\Delta^{(i)}_v(1+\beta)).
\end{equation}
Thus the dynamic PGD can lead to a finer search of a minimal budget by repeatedly crossing the decision boundary.

\textbf{Initialization of $\boldsymbol{\delta}_v$.} Here we point that the attack performance of our MiBTack may rely on the initial status $\boldsymbol{a}^{\prime(0)}_v = \boldsymbol{a}_v + \boldsymbol{\delta}^{(0)}_v = \boldsymbol{a}_v$ with $\boldsymbol{\delta}^{(0)}_v=\boldsymbol{0}$.
Since the existing iterative attacks often add noises monotonically along the direction of gradient~\cite{Shi2019CurlsW}, resulting in a dependence of initial direction. 
While in multi-classification, the initial attack direction may not be optimal: when minimize CW-based loss $\mathcal{L}(\boldsymbol{a}^{\prime(0)}_v)$, the yielded perturbation often move node to the wrong class $c$ which has the largest confidence ($\mathop{\rm max}\limits_{c \neq y_v}  f_{\boldsymbol{\theta}^*}(\boldsymbol{a}_v)_{c}$). We find that it often fails to indicate the closest decision boundary in multi-class classification. 
Clearly, precisely estimating the closest decision boundary beforehand and along its direction guiding the later attack generation can help generate a more unnoticeable attack. Here we solve this problem by performing topology attacks with one-step towards each wrong class, and choose the class with the largest descent of $\mathcal{L}$, then use it to initialize $\boldsymbol{\delta}_v$. The details can be found in Appendix~\ref{app:initial}.

\textbf{Convergence.}
We alternatively update perturbation and budget until target node reaches the decision boundary, then we are only allowed to iterate  for a given patience $P$ times. To accelerate convergence and dampen oscillations around the boundary in last $P$ iterations~\cite{Pintor2021FastMA}, we start to use cosine annealing to reduce the step sizes, including step size $\alpha$ for updating $\boldsymbol{\delta}^{(i-1)}_v$ in Eq.\ref{eq:PGD_gd} and $\beta$ for updating $\Delta^{(i)}_v$ in Eq.\ref{eq:dyn_shrink} and Eq.\ref{eq:dyn_enlarge}.  Finally, the perturbed target node will reach a point where the (green) decision boundary is tangent to the (dotted green) sphere of $\Delta^{\star}_v$ as shown in Fig.~\ref{fig:butterfly}, yielding the optimal $\boldsymbol{\delta}_v^{\star}$ with minimum budget $\Delta^{\star}_v$.
We summarize the pseudo-code in Appendix~\ref{app:alg} for the whole process of MiBTack. Moreover, the space consumption analysis is provided in Appendix~\ref{app:space}.

\section{Experiments}

\subsection{Experimental Setup}\label{sec:5.1}

\textbf{Datasets.} 
We employ the widely-used citation networks (Cora, Citeseer, Pubmed) as in~\cite{nettack}, and a social network Polblogs in~\cite{polblogs}. We conduct the experiments with 250 randomly selected nodes in the test set as the targeted nodes that are to be attacked following~\cite{DBLP:conf/icdm/WangLYL0FH20}. More details of dataset are shown in Appendix~\ref{app:data}.

\begin{table*}[!htbp]
\caption{Attack performance. The lower classification accuracy (ACC) of all attacked target nodes indicates a better attack performance. The lower total budget (TB) hints the better invisibility.}
\vspace{-5pt}
\resizebox{0.85\textwidth}{!}{
\begin{tabular}{|c|c|c|cc|ccccc|c|}
\hline
Datasets  & GNNs  & Metrics &{PGD} &{PRBCD} & {Rand}  & {DICE} & {DICE-t}  &{FGA} & {Nettack}  & {MiBTack} \\ \hline
  &  & ACC &0.032	&0.020	&0.112	&0.000 &0.052	&0.000	&0.000	&\textbf{0.000} \\  
  & \multirow{-2}{*}{GCN} & TB &765	&779	&10349	&4057 &1171	&384	&357	&\textbf{330} \\ \cline{2-11} 
  &  & ACC &0.020	&0.008	&0.104	&0.000 &0.000	&0.000	&0.000	&\textbf{0.000} \\  
  & \multirow{-2}{*}{SGC} & TB &787	&792	&6774	&4927 &1155	&437	&443	&\textbf{385} \\ \cline{2-11} 
  &  & ACC &0.256	&0.100	&0.152 &0.000 &0.000	&0.000	&0.000	&\textbf{0.000} \\  
\multirow{-6}{*}{Cora} & \multirow{-2}{*}{APPNP} & TB &777	&796	&2919	&2518 & 1237	&683	&679	&\textbf{507} \\ \hline
  &  & ACC &0.100	&0.092	&0.132	&0.000 &0.000	&0.000	&0.000	&\textbf{0.000} \\  
  & \multirow{-2}{*}{GCN} & TB &691	&697	&4290	&4292 &1343	&484	&\textbf{438}	&444 \\ \cline{2-11} 
  &  & ACC &0.048	&0.028	&0.144	&0.000 &0.000	&0.000	&0.000	&\textbf{0.000} \\  
  & \multirow{-2}{*}{SGC} & TB &689	&693	&6362	&3461 &1002	&411	&414	&\textbf{386} \\ \cline{2-11} 
  &  & ACC &0.172	&0.136	&0.064	&0.000 &0.016	&0.000	&0.000	&\textbf{0.000} \\  
\multirow{-6}{*}{Citeseer} & \multirow{-2}{*}{APPNP} & TB &650	&696	&7386	&2201 &949	&494	&551	&\textbf{412} \\ \hline
  &  & ACC &0.000	&0.004	&0.380	&0.000 &0.000	&0.000	&0.000	&\textbf{0.000} \\  
  & \multirow{-2}{*}{GCN} & TB &7112	&7105	&18891 	&7261 &7261	&2038	&2011	&\textbf{1961} \\ \cline{2-11} 
  &  & ACC &0.036	&0.064	&0.468	&0.000	&0.000 &0.000	&0.000	&\textbf{0.000} \\  
  & \multirow{-2}{*}{SGC} & TB &7027	&7062	&19250 &7304	&7304	&3518	&3660	&\textbf{3328} \\ \cline{2-11} 
  &  & ACC &0.396	&0.132	&0.528 &0.000	&0.000	&0.000	&0.000	&\textbf{0.000} \\  
\multirow{-6}{*}{Polblogs} & \multirow{-2}{*}{APPNP} & TB &7125	&7125	&14551 &7543	&7543	&5619	&5739	&\textbf{5494} \\ \hline
  &  & ACC &0.108	&0.016	&0.292 &0.000	&0.000	&0.000	&0.000	&\textbf{0.000} \\  
  & \multirow{-2}{*}{GCN} & TB &742	&770	&2935 	&4857 &1386	&357	&354	&\textbf{348} \\ \cline{2-11} 
  &  & ACC &0.196	&0.020	&0.284 &0.000	&0.000	&0.000	&0.000	&\textbf{0.000} \\  
  & \multirow{-2}{*}{SGC} & TB &683	&772	&6740	&2540 &1346	&368	&364	&\textbf{359} \\ \cline{2-11} 
  &  & ACC &0.296	&0.148	&0.284 &0.000	&0.000	&0.000	&0.000	&\textbf{0.000} \\  
\multirow{-6}{*}{Pubmed} & \multirow{-2}{*}{APPNP} & TB &720	&744	&2111 	&2340 &1241	&529	&\textbf{514}	&527 \\ \hline
\end{tabular}
}
\label{tab:attack_performance}
\end{table*}

\textbf{Baselines.}
Since existing topology attack methods are all fixed-budget attacks, here we adapt the state-of-the-art baselines to minimum-budget topology attack. For greedy-based attacks, \textbf{Rand} randomly flips the edges in $\mathbf{a}_v$ for target node $v$. 
\textbf{DICE}~\cite{Waniek2017HidingIA} randomly disconnects the links of $v$ to the neighbors with the same label $y_v$ and connects the $v$ to the nodes with class $c \neq y_v$. 
\textbf{DICE-t} extends DICE by adding the edges $(v,u)$ where $u$ belongs to the target class $c^*$ and $c^* = \mathop{\rm arg max}\limits_{c \neq y_v} f_{\boldsymbol{\theta}^*}(\boldsymbol{a}_v)$.
\textbf{FGA}~\cite{geisler2021robustness} flips one edge at a time by performing gradient update along the direction of the sign of gradients of loss function w.r.t. each adjacency matrix.
\textbf{Nettack}~\cite{nettack} generates perturbation edges greedily by exploiting the properties of the linearized GCN surrogate.
We can easily adapt above greedy-based attacks by consistently adding perturbations until target node $v$ is misclassified or the number of perturbed edges is more than 1000.
For PGD-based attacks, \textbf{PGD}~\cite{Xu2019TopologyAA} uses projected gradient descent to project the perturbation edges into the space of given budget $\Delta$.
\textbf{PRBCD}~\cite{geisler2021robustness} generates $\Delta$ perturbation edges for large-scale graph by projected randomized block coordinate descent. Above PGD-based algorithms model the fixed budget attacks as a convex constrained optimization, and cannot be directly used for our minimum-budget attacks which is essentially a non-convex constrained optimization problem. Following~\cite{geisler2021robustness}, we use the degree of the target node that we currently attack as budget. 
More implementation details of baselines and our MiBTack, e.g., hyperparameters, are provided in Appendix~\ref{app:codes} and~\ref{app:exp_setting}.

\textbf{Target Models.} To validate the generalization ability of our proposed attacker, we choose three popular graph neural networks~\footnote{The code of GNNs can be found in \url{https://github.com/BUPT-GAMMA/GammaGL}.}:
(1) GCN~\cite{GCN} is a representative GNN and learns on graph structures using convolution operations. We train a 2-layer GCN with learning rate 0.01, where the number of units in hidden layer is 16. In addition, the dropout rate is 0.5, weight decay is $5e\text{-}4$. 
(2) SGC~\cite{Wu2019SimplifyingGC} simplifies GCNs through successively removing nonlinearities and collapsing weight matrices between consecutive layers.  For SGC, the learning rate is 0.01, the number of units is 16,  the dropout rate is 0.5, and weight decay is $5e\text{-}6$. 
(3) APPNP~\cite{Klicpera2019PredictTP} improves GCN by leveraging residual connection to preserve the information of raw features. The learning rate of APPNP is 0.01, the number of units in hidden layer is 64, the dropout rate is 0.5, weight decay is $5e\text{-}6$. 

% To validate the generalization ability of our proposed attacker, we choose three popular graph neural networks: GCN~\cite{GCN}, SGC~\cite{Wu2019SimplifyingGC} and APPNP~\cite{Klicpera2019PredictTP}. The detailed information can be found in Appendix~\ref{app:gnns}.

% \textbf{Hyperparameters.}

\textbf{Metrics.}
Here we utilize two metrics to evaluate the minimum-budget topology attacks. (1) Accuracy ({ACC}): We use the accuracy of GNN model on all attacked target nodes to show whether the attack models can make all nodes misclassification. The zero value of ACC indicates 100\% attack successful rate.
(2) Total Budget ({TB}): Total budgets is the amount of perturbation edges for attacking all target nodes. 
A lower TB hints a more unnoticeable attack.

\subsection{Attack Effectiveness}
Here we evaluate the effectiveness of our model against all baselines for minimum-budget topology attacks, under four datasets and three GNNs. The overall results are presented in Table~\ref{tab:attack_performance}, where we have the following observations:\\
$\bullet$ Our MiBTack can outperform all baselines in most scenarios, yielding minimum attacks with the guarantee of misclassification of all nodes. 
First, the PGD-based attacks {PGD} and {PRBCD} are hard to achieve 0 accuracy and need large budgets, since they predefine the node budget by its degree, which is usually not optimal. 
Based on random strategies, {Rand}, {DICE} and {DICE-t}, need large amounts of perturbation edges. Then Nettack and FGA have better performance than above methods, since they greedily flip the adversarial edges with the highest gradient w.r.t. attack loss. Finally, our MiBTack can generate the minimum attacks with the guarantee of 0 accuracy, averagely saving at least 60 perturbed edges. Compared to myopic greedy search, MiBTack can search the better combinatorial perturbed edges based on our dynamic PGD. \\
$\bullet$ Specifically, we observe that our MiBTack has the largest improvement on Polblogs. This is because that the topology connections in Polblogs are much more dense than other datasets, thus the nodes in Polblogs require significantly larger budget to successfully attack, namely more iterations, where greedy based baselines will accumulate more errors than gradient based methods. \\
$\bullet$ We also observe that, under most scenarios, APPNP needs higher TB than GCN and SGC, indicating more adversarial robustness of APPNP. The reason may be that APPNP, which leverages residual connection to preserve the information of raw features, may be less dependent on topology and thus have better robustness to topology attacks. Besides, GNNs are more robust on Polblogs due to the relatively more dense graph of Polblogs.

\begin{figure}[!htbp]
\minipage{0.23\textwidth}
\centering
 \subfigure[GCN.]{\includegraphics[ width=\textwidth]{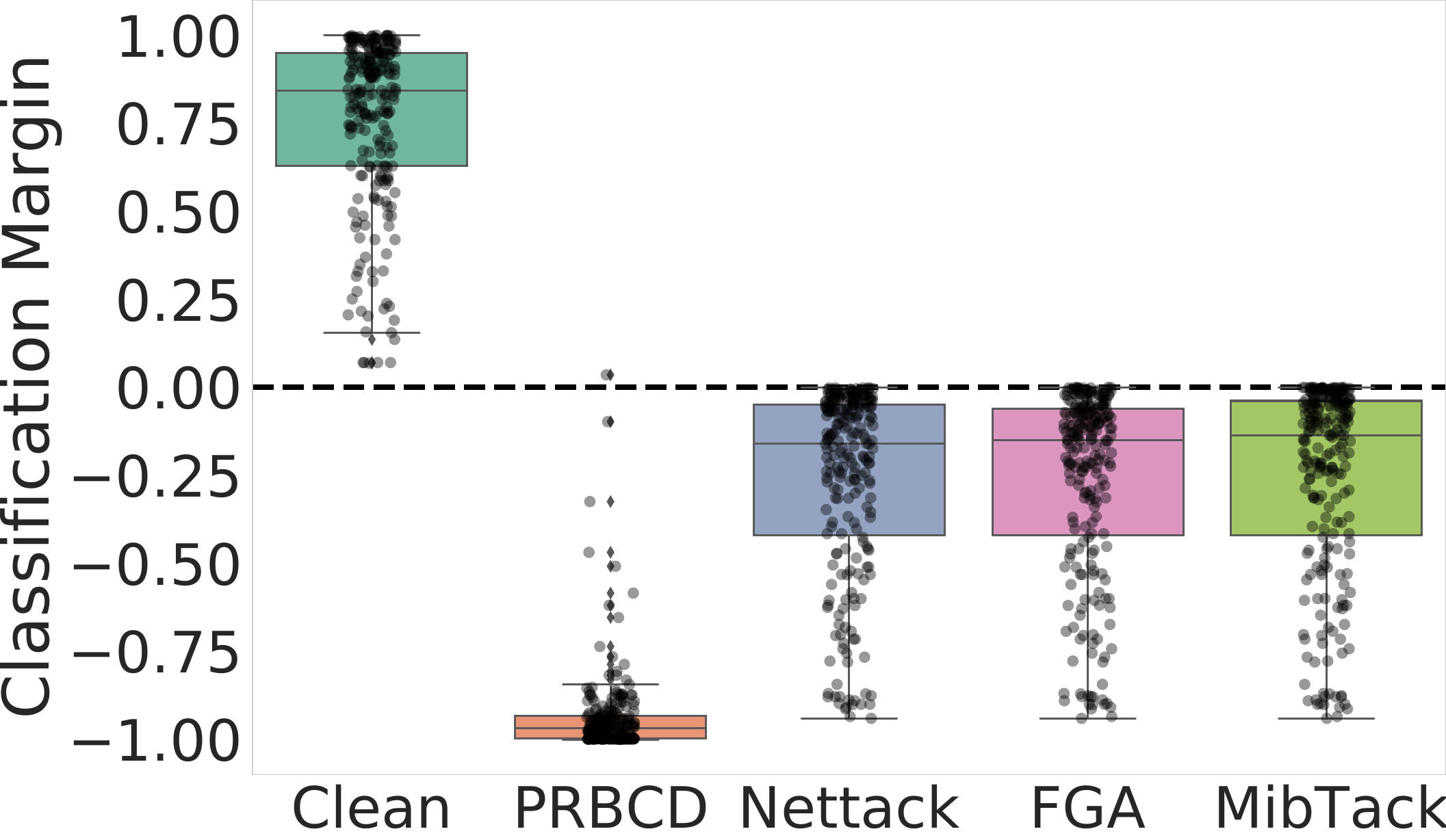}}
\endminipage\hfill
\minipage{0.23\textwidth}
\centering
 \subfigure[APPNP.]{\includegraphics[ width=\textwidth]{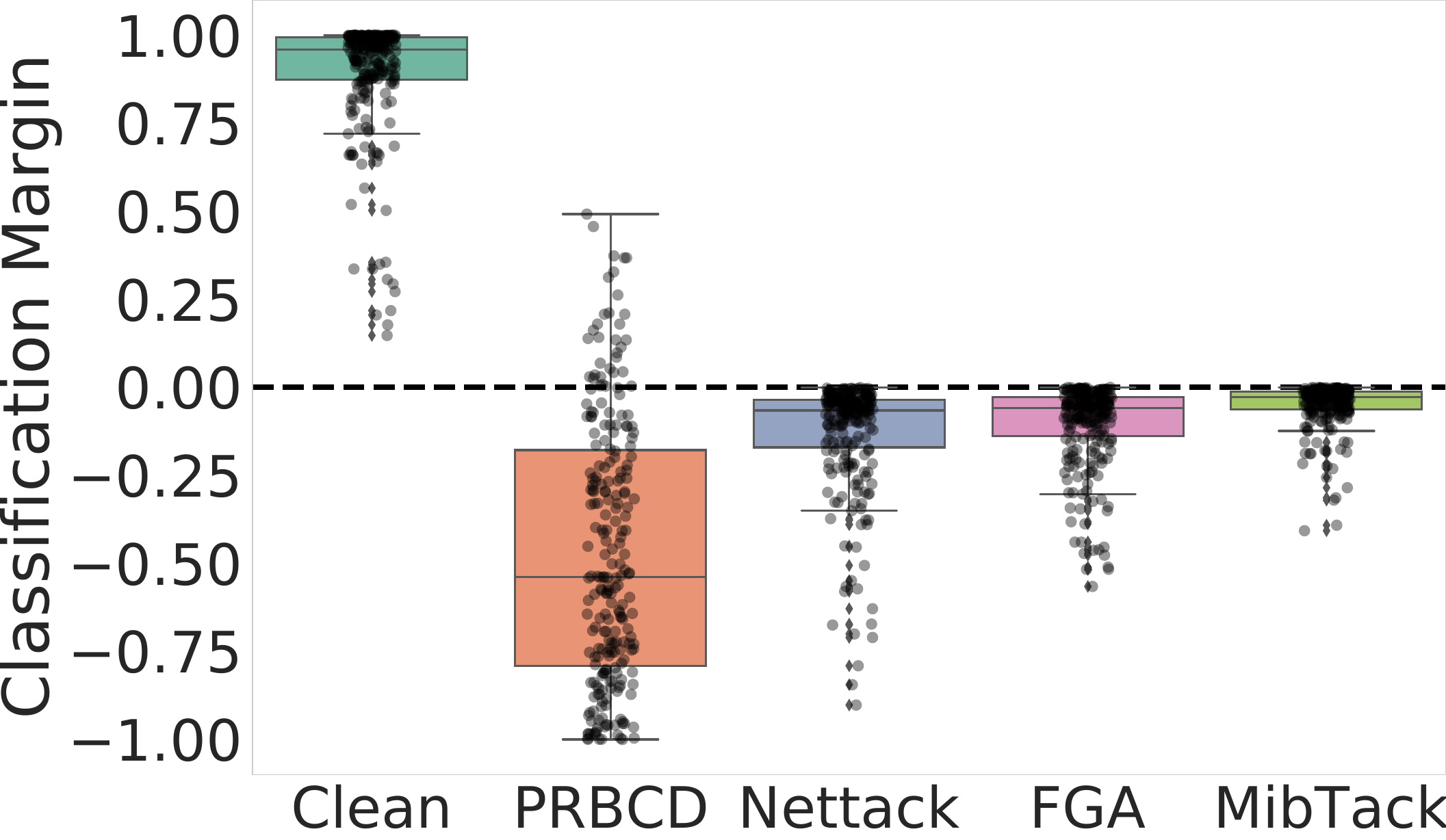}}
\endminipage\hfill
\vspace{-10pt}
\caption{The node classification margins on Polblogs. Our MiBTack leads all target nodes misclassified and affects them most slightly by finding minimum and successful attacks.}
\label{fig:margin}
\vspace{-5pt}
\end{figure}

To better evaluate how effective is our model, take the more robust Polblogs dataset as an example, we  plot the classification margins of GCN and APPNP in Fig.~\ref{fig:margin}. Each point in the plot represents one target node $v$. The classification margin of $v$ is $f_{\boldsymbol{\theta}}(\boldsymbol{a}_{v})_{y_v} - \max _{c \neq y_v} f_{\boldsymbol{\theta}}(\boldsymbol{a}_{v})_{c}$ where $y_v$ is the ground truth class, $f_{\boldsymbol{\theta}}(\boldsymbol{a}_{v})_{y_v}$ is the probability of node $v$ being classified into class $y_v$.  The positive classification margin value of node $v$ indicates a failed attack, where $v$ fails to cross the decision boundary. For the negative value, $v$ is misclassified and a lower value indicates that $v$ crosses the decision boundary more. Compared to the clean scenario, we find that PRBCD with node degree as budget, can strongly affect GNNs most, but fail to lead the misclassification for all target nodes. On the contrast, our MiBTack and the greedy-based attacks, FGA and Nettack, affect more slightly but successfully attack all target nodes. Most remarkably, our MiBTack achieves higher margin value than FGA and Nettack, indicating that the nodes, attacked by MiBTack, cross the decision boundary less. This is because that the dynamic PGD in our MiBTack can converge to decision boundary and lead to a finer search of a minimal  combination of perturbation edges by repeatedly crossing the decision boundary.

\begin{figure*}[!tbp]
\minipage{0.24\textwidth}
\centering
 \subfigure[Cora, ACC.]{\includegraphics[ width=\textwidth]{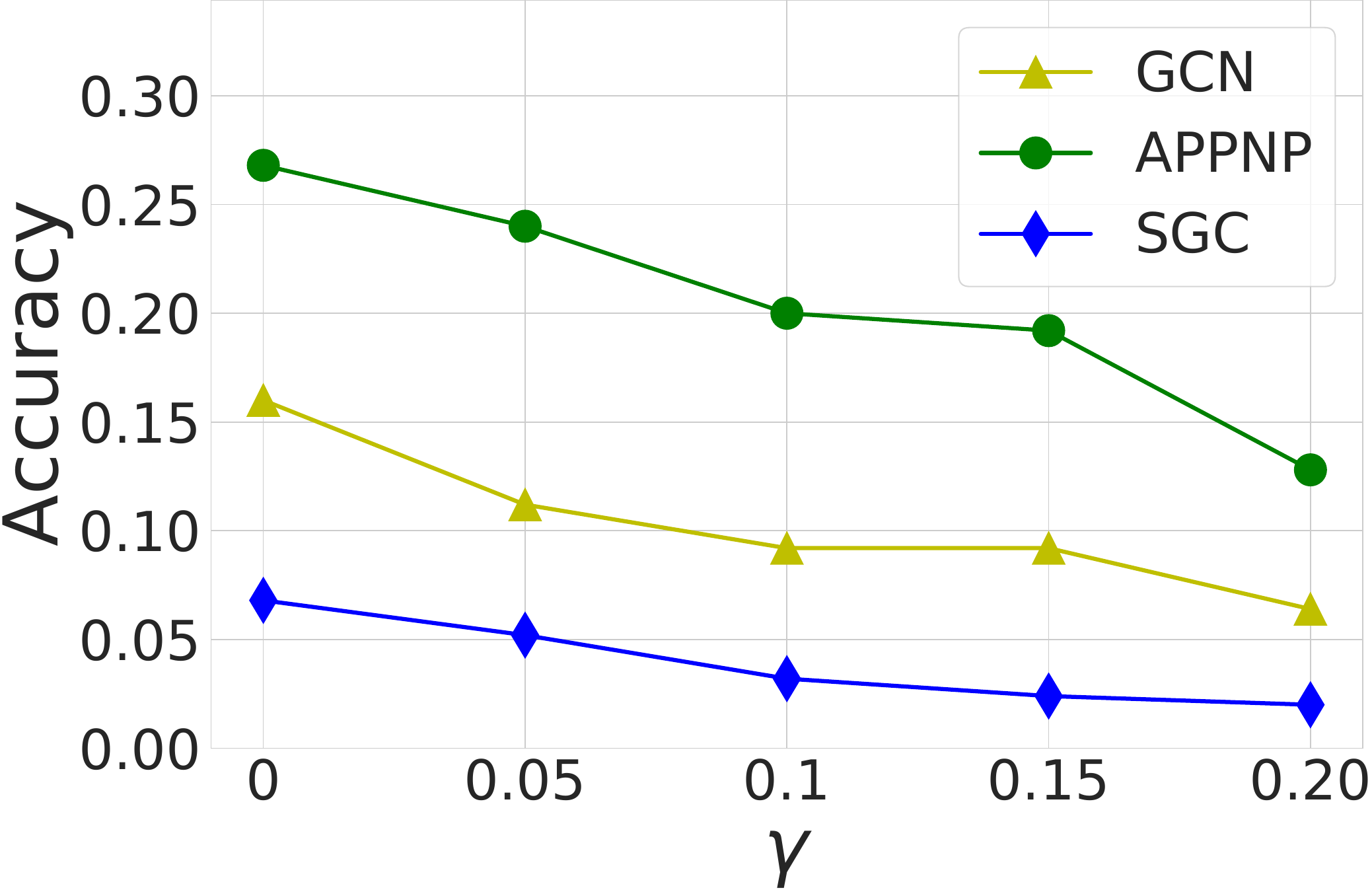}}
\endminipage\hfill
\minipage{0.24\textwidth}
\centering
\subfigure[Citeseer, ACC.]{\includegraphics[ width=\textwidth]{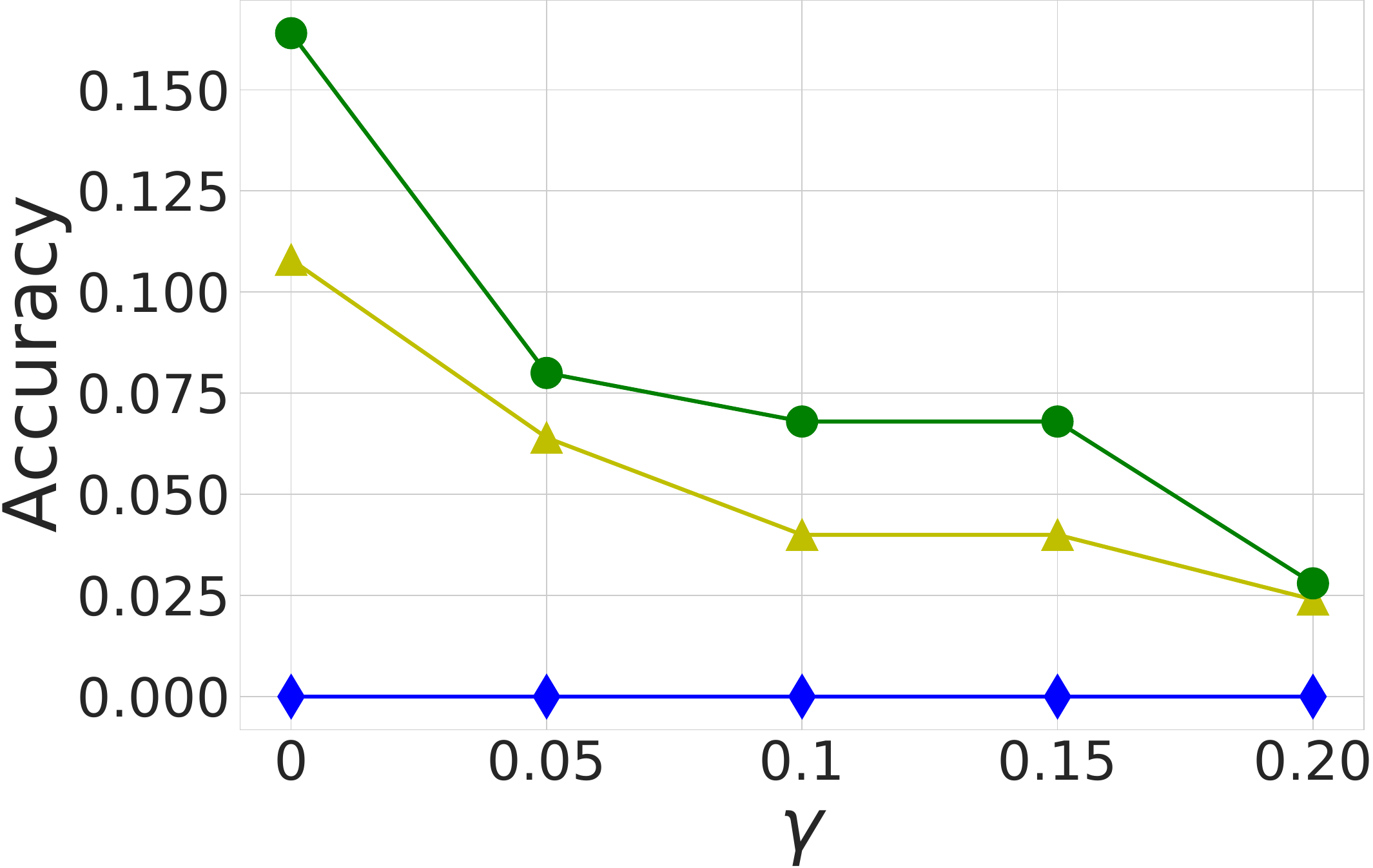}}
\endminipage\hfill
\minipage{0.24\textwidth}
\centering
 \subfigure[Cora, Total Budget.]{\includegraphics[ width=\textwidth]{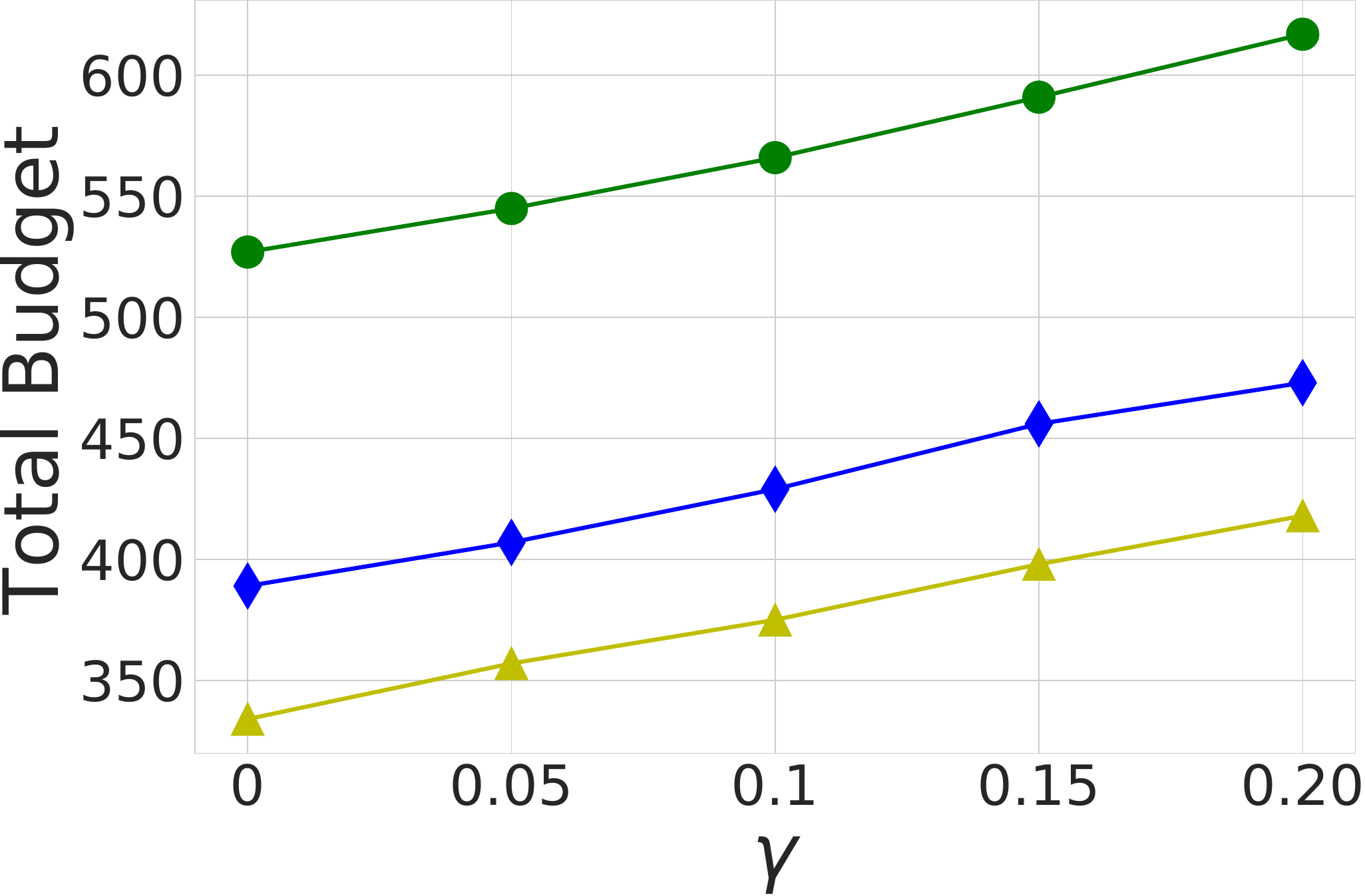}}
\endminipage\hfill
\minipage{0.24\textwidth}
\centering
 \subfigure[Citeseer, Total Budget.]{ \includegraphics[ width=\textwidth]{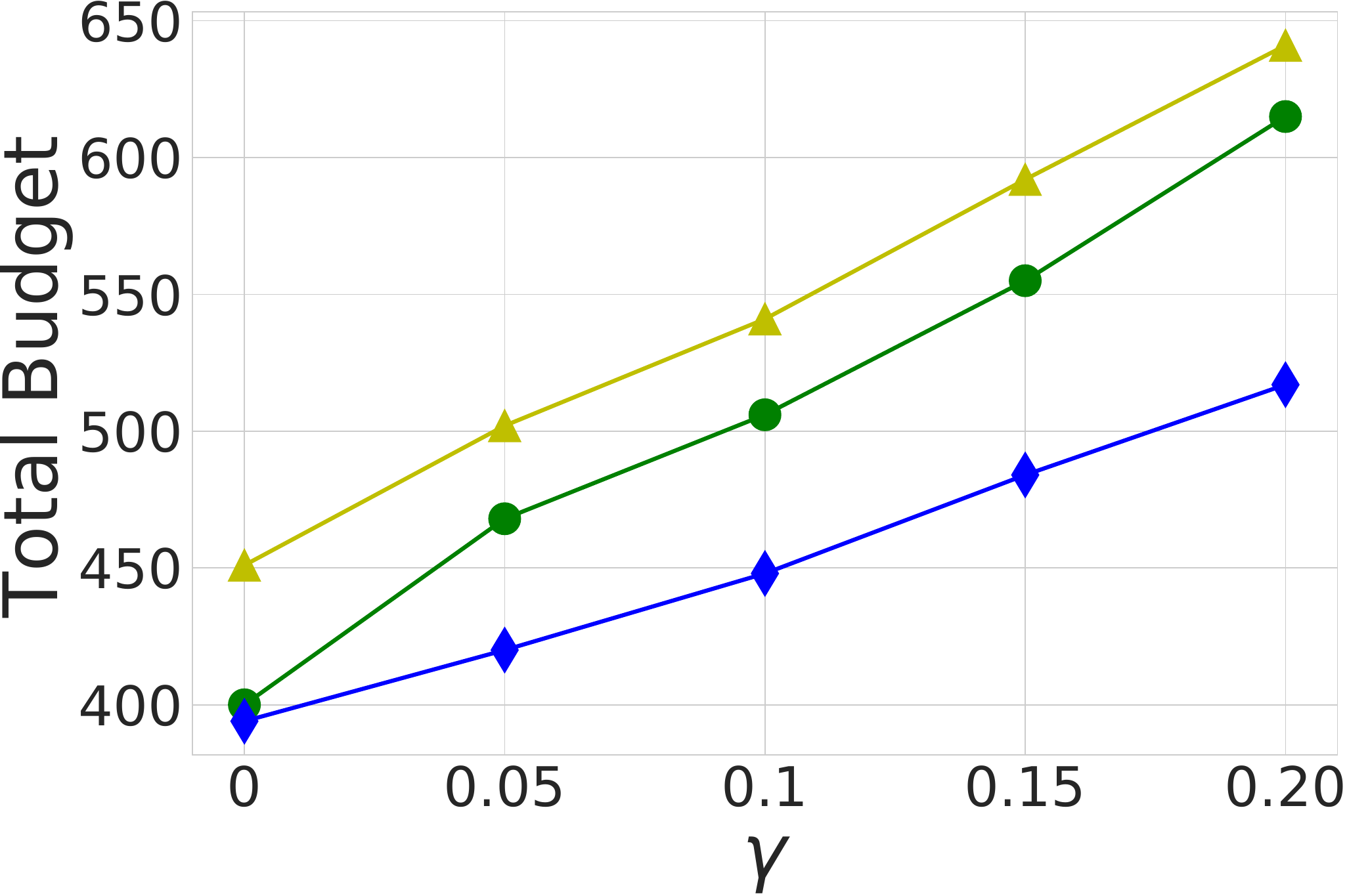}}
\endminipage\hfill
\vspace{-10pt}
\caption{Performance of poisoning attacks on Cora and Citeseer. 
}
\label{fig:poison}
\vspace{+10pt}
\minipage{0.24\textwidth}
\centering
 \subfigure[Cora, JacGCN.]{\includegraphics[ width=\textwidth]{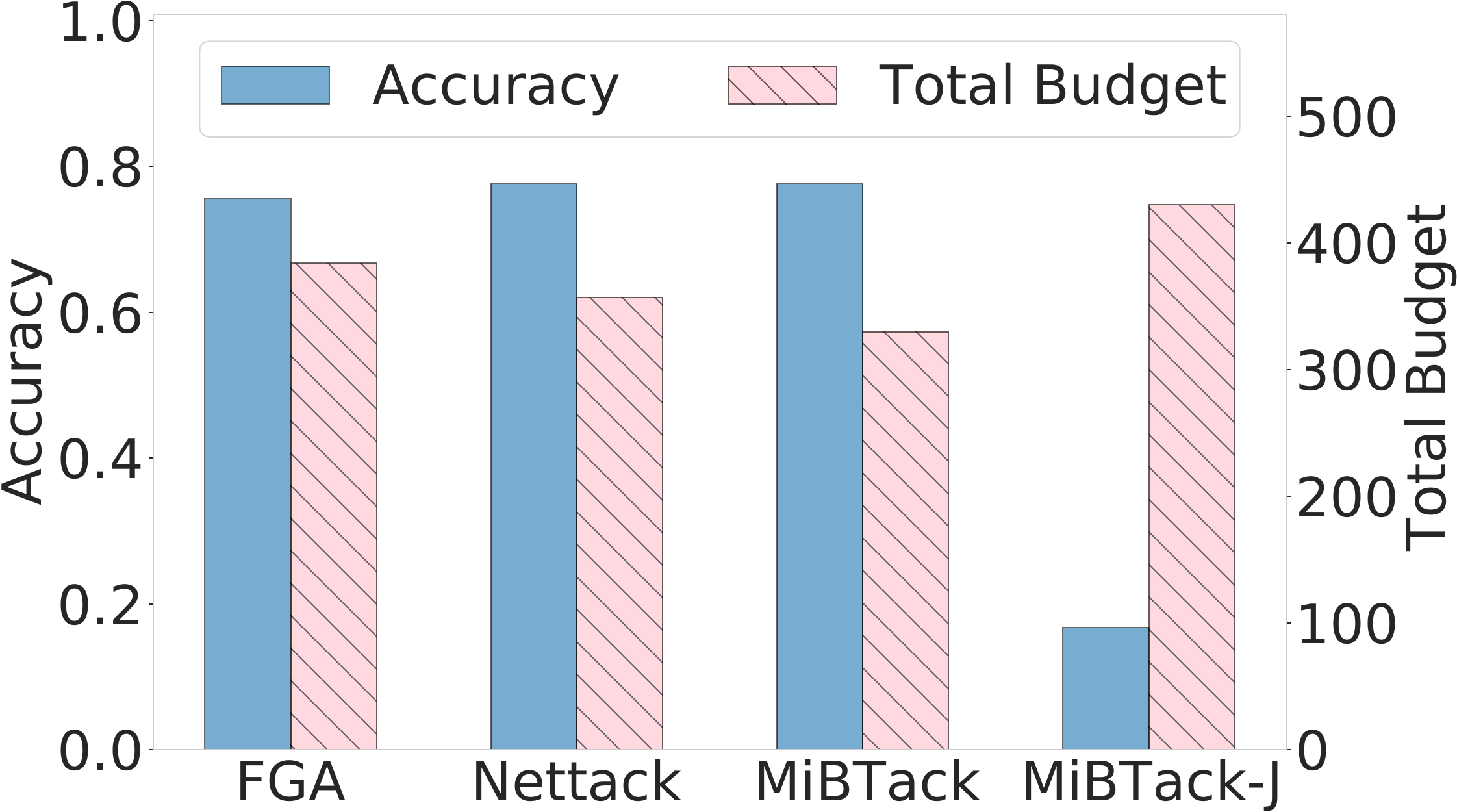}}
\endminipage\hfill
\minipage{0.24\textwidth}
\centering
 \subfigure[Citeseer, JacGCN.]{\includegraphics[ width=\textwidth]{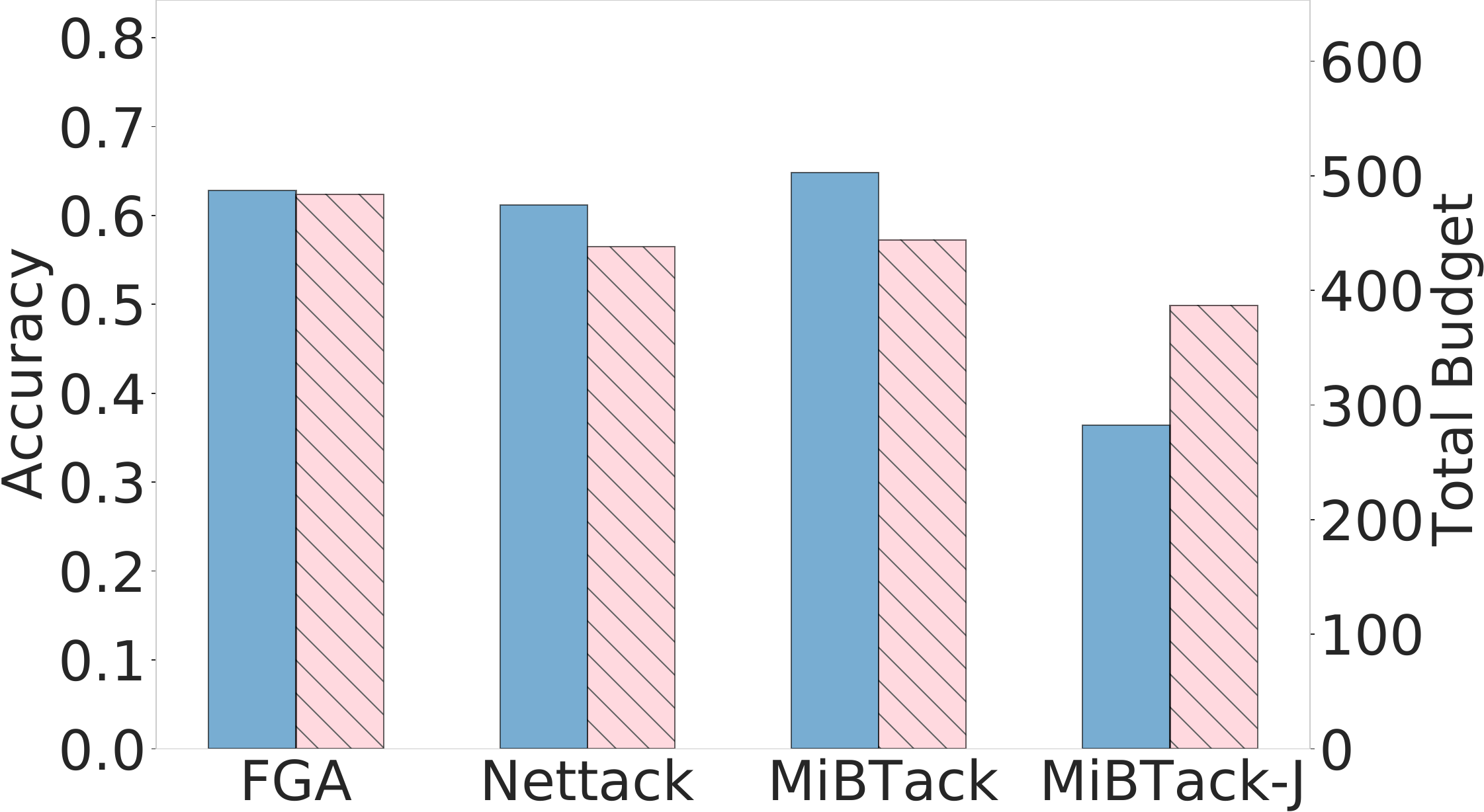}}
\endminipage\hfill
\minipage{0.24\textwidth}
\centering
\subfigure[Cora, RGCN.]{\includegraphics[ width=\textwidth]{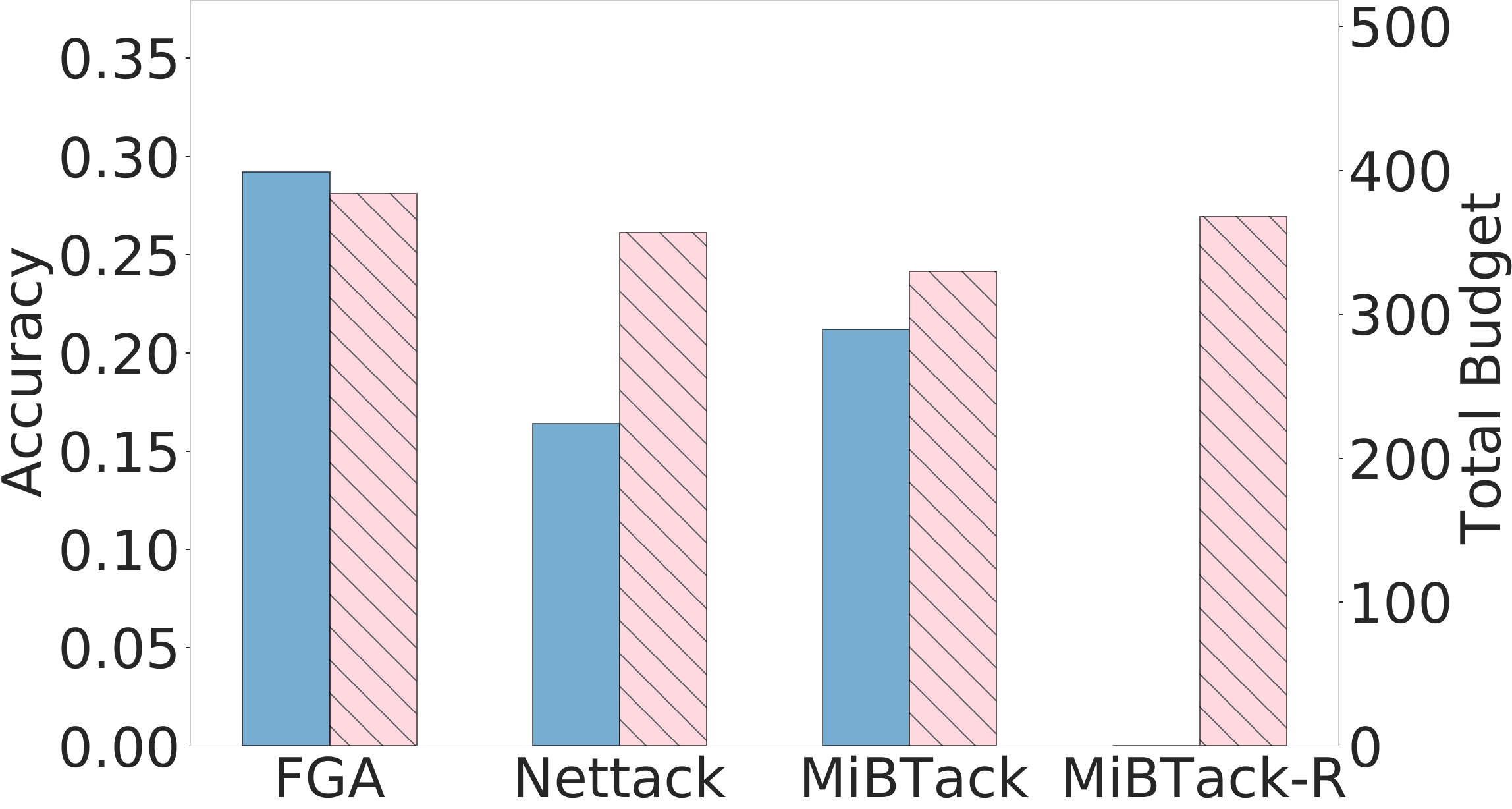}}
\endminipage\hfill
\minipage{0.24\textwidth}
\centering
 \subfigure[Citeseer, RGCN.]{ \includegraphics[ width=\textwidth]{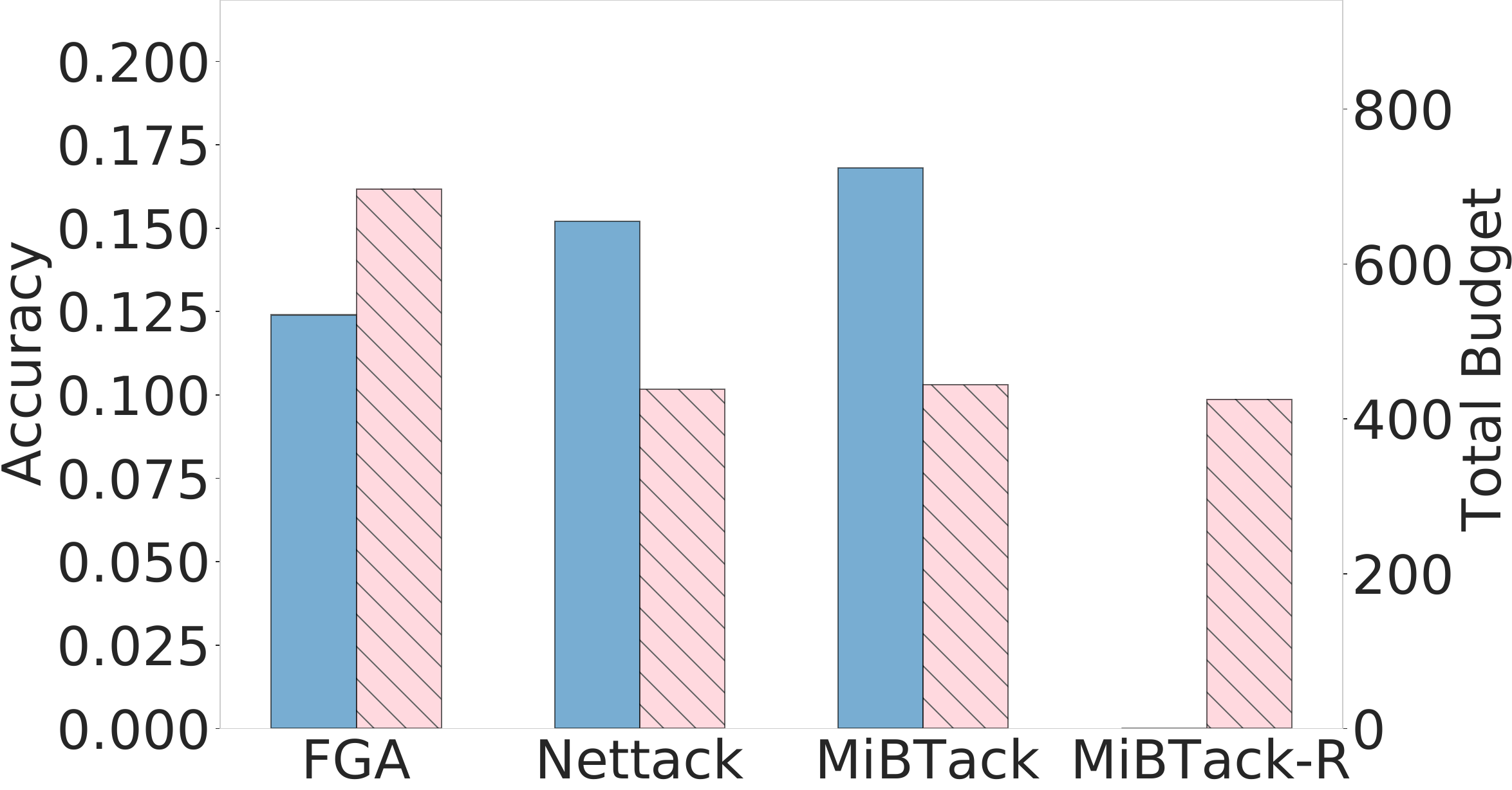}}
\endminipage\hfill
\vspace{-10pt}
\caption{Performance of attacks against defensed GNNs (i.e., JacGCN and RGCN). We report the total budget and the accuracy of target nodes of defensed GNNs under FGA, Nettack, our MiBTack, our variants MiBTack-J and MiBTack-R.}
\label{fig:defense}
\end{figure*}

\subsection{Transferability of Attacks} 
We mainly focus on evasion (test time) attacks against GNNs, but we also evaluate the transferability of our MiBTack under poisoning (train time) setting and defensed setting. 

\textbf{Poisoning Attacks}.  Following~\cite{geisler2021robustness}, we retrain GNNs on the perturbed graph of an evasion attack for poisoning.
Note that poisoning attack is a more challenging scenario since retraining GNNs will introduce much uncertainty to attackers. Fortunately, our MiBTack can handle such uncertainty by improving the confidence level of misclassification $\gamma$.  
Take Cora and Citeseer as examples, we gradually increase $\gamma$ and present the ACC of attacked target nodes in Fig.~\ref{fig:poison} (a)  and (b). As seen, for each GNN model, the accuracy of target nodes is close to 0 and shows the similar declining trend with the increasing $\gamma \in \{0.00, 0.05, 0.10, 0.15, 0.20\}$. So MiBTack with larger $\gamma$ tends to generate stronger attack under poisoning scenarios.  But there is no free lunch: %we have to expect that the needed total budget will grow, and indeed, %this is so 
it is expected that the needed total budget will grow, as showed in Fig.~\ref{fig:poison} (c) and (d). In conclusion, our MiBTack can achieve powerful attacks under challenging poisoning attacks. 

\textbf{Attacks against Defensed GNNs}.
Some countermeasures are proposed to improve the robustness of GNNs~\cite{Li2022ReliableRM,Liu2022ANGCNAA,Zhuang2022DefendingGC,Zhang2022GraphAL,Chang2021NotAL,Liu2021GraphNN,Xu2021SpeedupRG,Liu2021ElasticGN,Chen2021UnderstandingSV}. Besides vanilla GNNs,  we also consider these defensed GNNs: \textbf{RGCN}~\cite{Zhu2019RobustGC} adopts a variance-based attention mechanism to remedy the propagation of adversarial attacks.  \textbf{JacGCN}~\cite{Wu2019AdversarialEF} filters edges based on attribute Jaccard similarity (here threshold is 0.03).  
With no surprise, directly transferring attacks from vanilla GNNs to defensed GNNs will lead to a huge drop of attack performance as shown in Fig.~\ref{fig:defense}. The accuracy of target nodes attacked by Nettack, FGA and our MiBTack significantly increase, e.g., from accuracy 0 of GCN to around 0.7 of JacGCN on Cora dataset.  However, we can easily extend MiBTack against defensed GNNs. For JacGCN, we restrict the search space with Jaccard similarity when generating perturbation edges, yielding \textbf{MiBTack-J}. For RGCN, we  replace the target GNN in the attack model by RGCN, named \textbf{MiBTack-R}.
As seen, the accuracy of JacGCN and RGCN (i.e., blue bars) under MiBTack-J and MiBTack-R  dramatically drop even reach 0, costing only a little bit more budgets (i.e., pink bars). 
Doubtlessly, if the defensed GNNs are complex even black-box, the gradient of adjacency matrix is unavailable, and the straight solution in MiBTack-J or MiBTack-R is not feasible. 
We leave designing effective attack models against black-box GNNs as our future work.

\begin{figure*}[!htbp]
\minipage{0.49\textwidth}
\centering
\minipage{0.49\textwidth}
\centering
\subfigure[Cora.] {    \includegraphics[width=\textwidth]{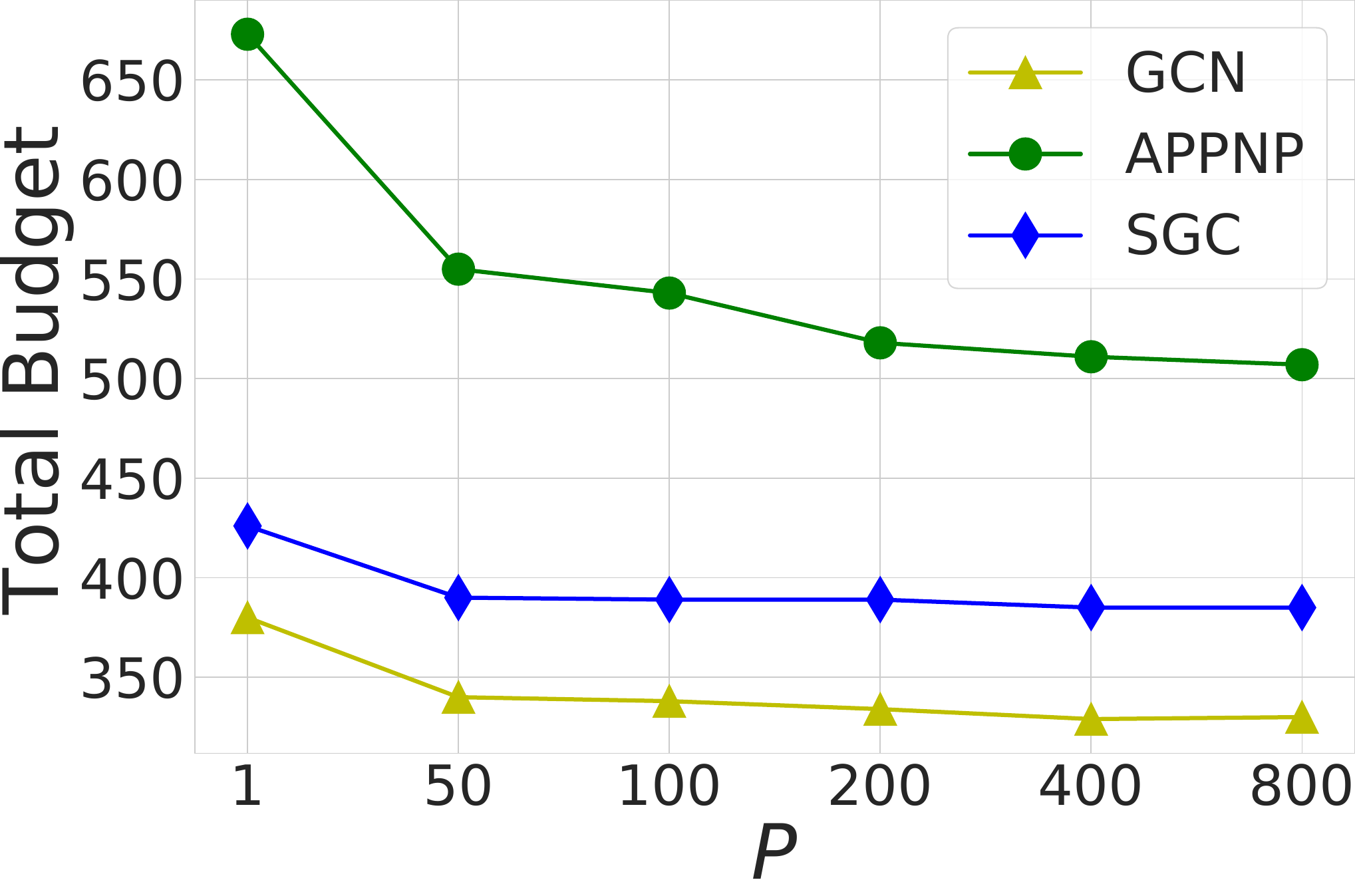}}
\endminipage\hfill
\minipage{0.49\textwidth}
\centering
\subfigure[Citeseer.] {    \includegraphics[width=\textwidth]{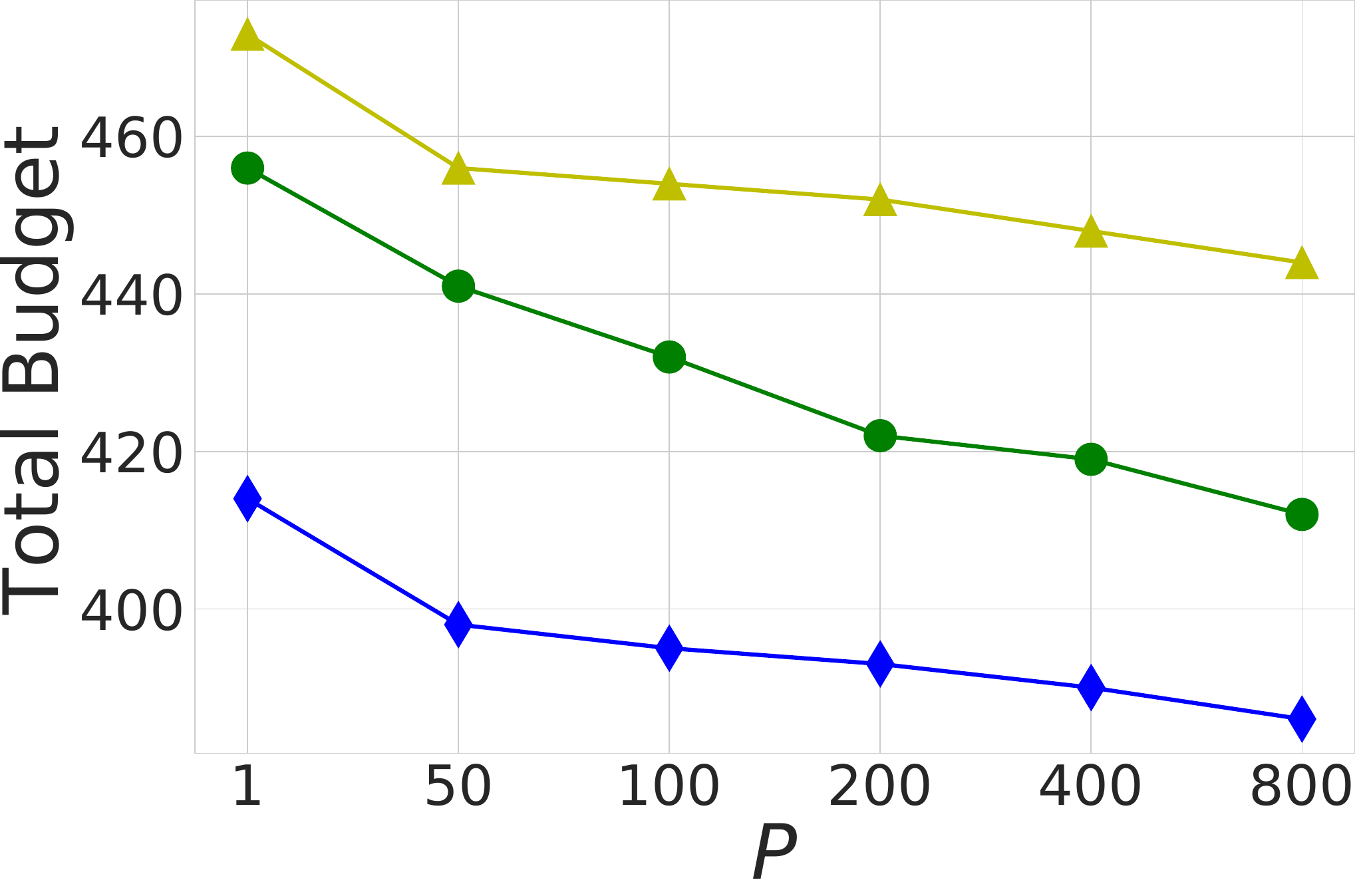}}
\endminipage\hfill
\vspace{-10pt}
\caption{Impact of $P$.}\label{fig:hyper_P}
\endminipage\hfill
\minipage{0.49\textwidth}
\centering
\minipage{0.49\textwidth}
\centering
\subfigure[Cora.] {    \includegraphics[width=\textwidth]{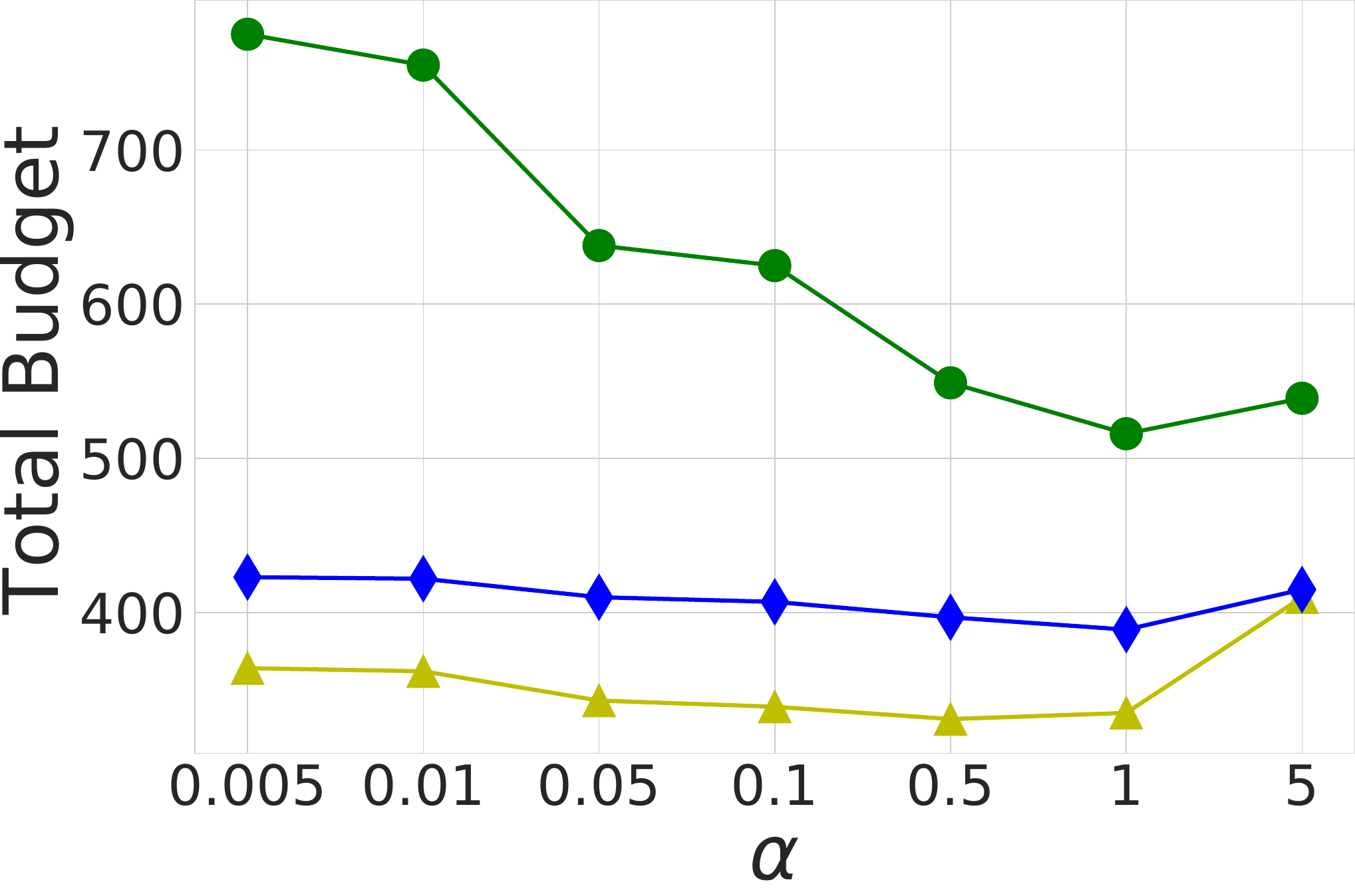}}
\endminipage\hfill
\minipage{0.49\textwidth}
\centering
\subfigure[Citeseer.] {    \includegraphics[width=\textwidth]{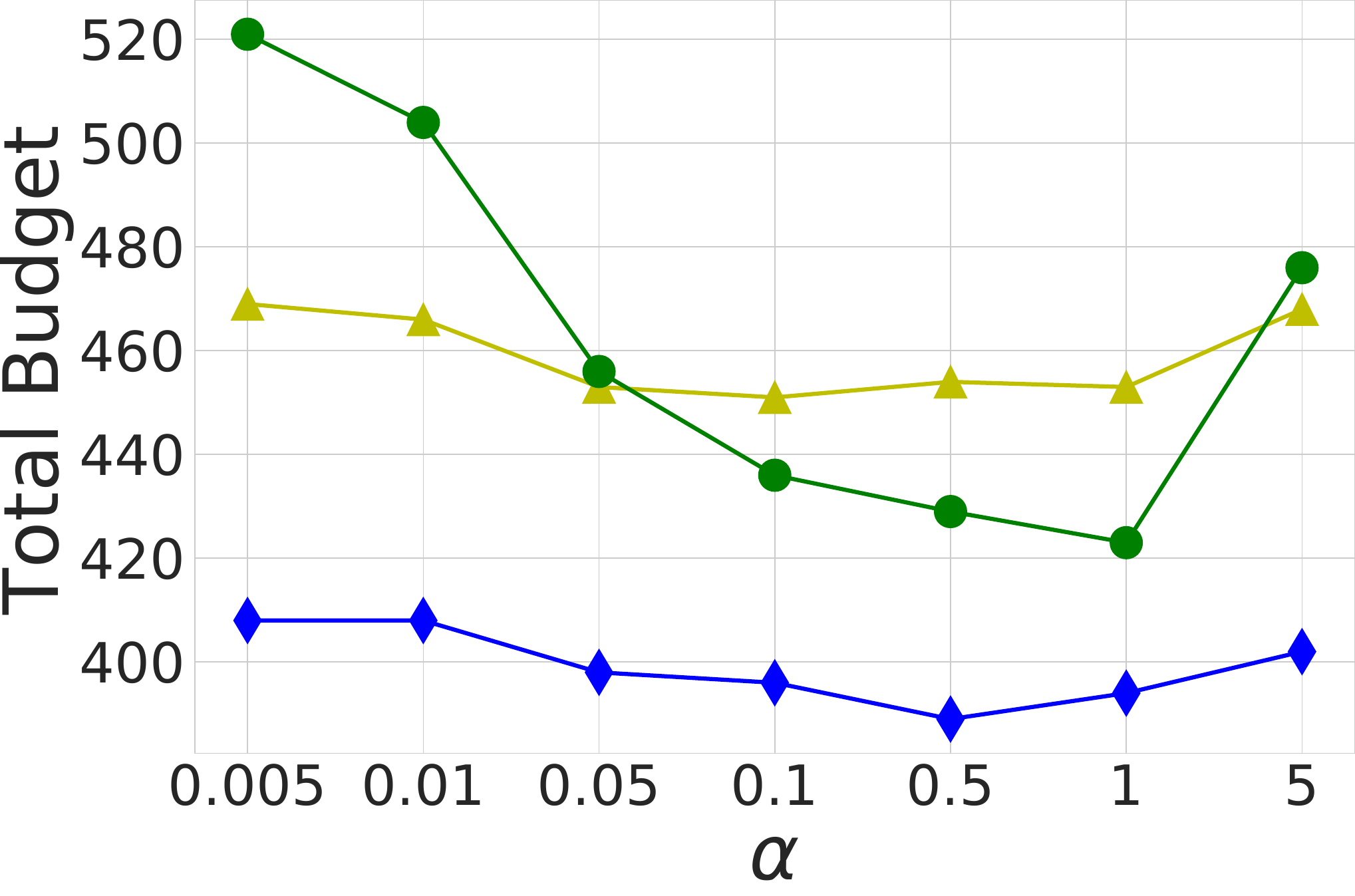}}
\endminipage\hfill
\vspace{-10pt}
\caption{Impact of $\alpha$.}\label{fig:hyper_alpha}
\endminipage\hfill
\end{figure*}

\begin{figure}[!htbp]
\minipage{0.49\textwidth}
\centering
\minipage{0.49\textwidth}
\centering
\subfigure[Cora.] {    \includegraphics[width=\textwidth]{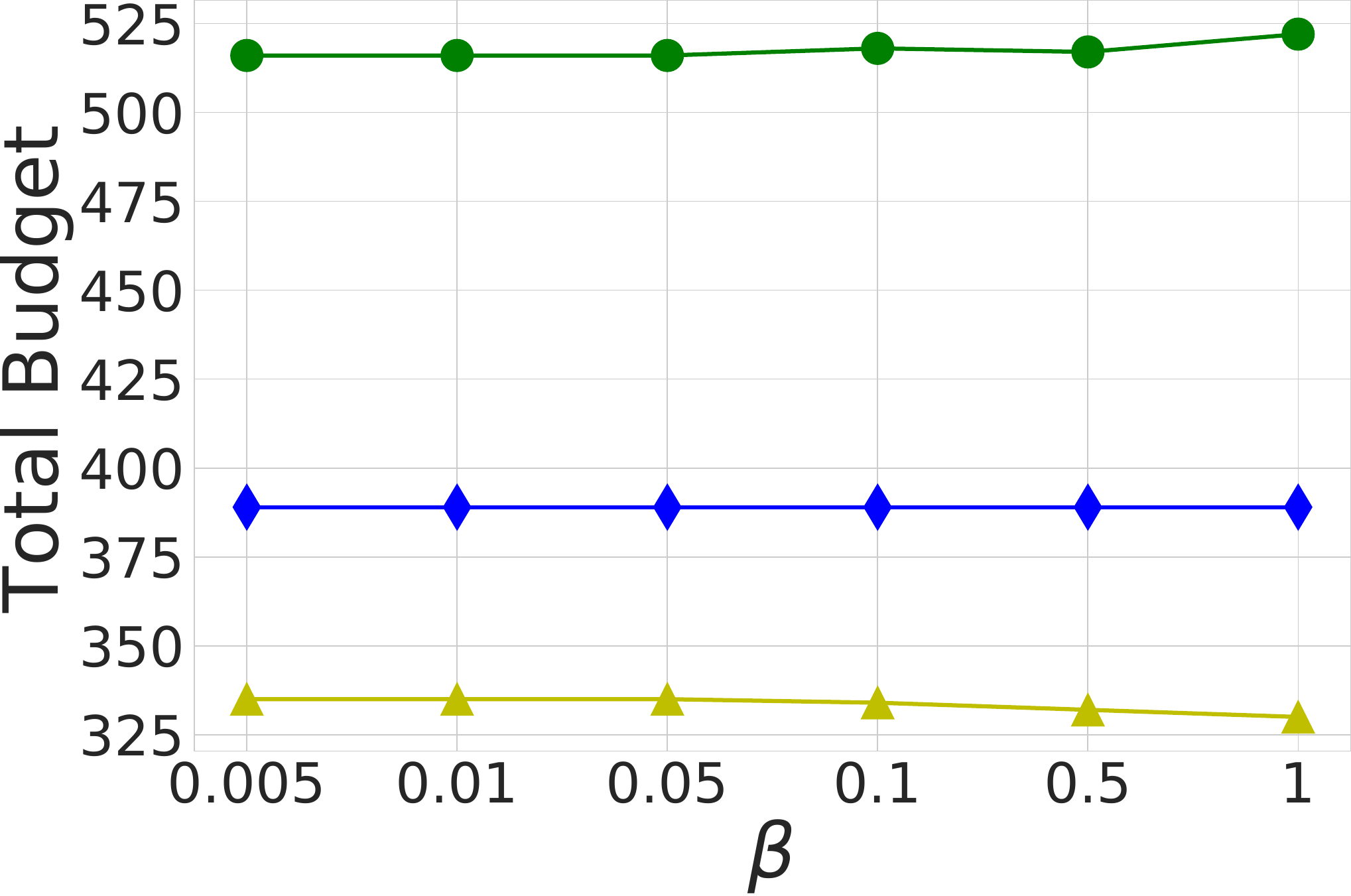}}
\endminipage\hfill
\minipage{0.49\textwidth}
\centering
\subfigure[Citeseer.] {    \includegraphics[width=\textwidth]{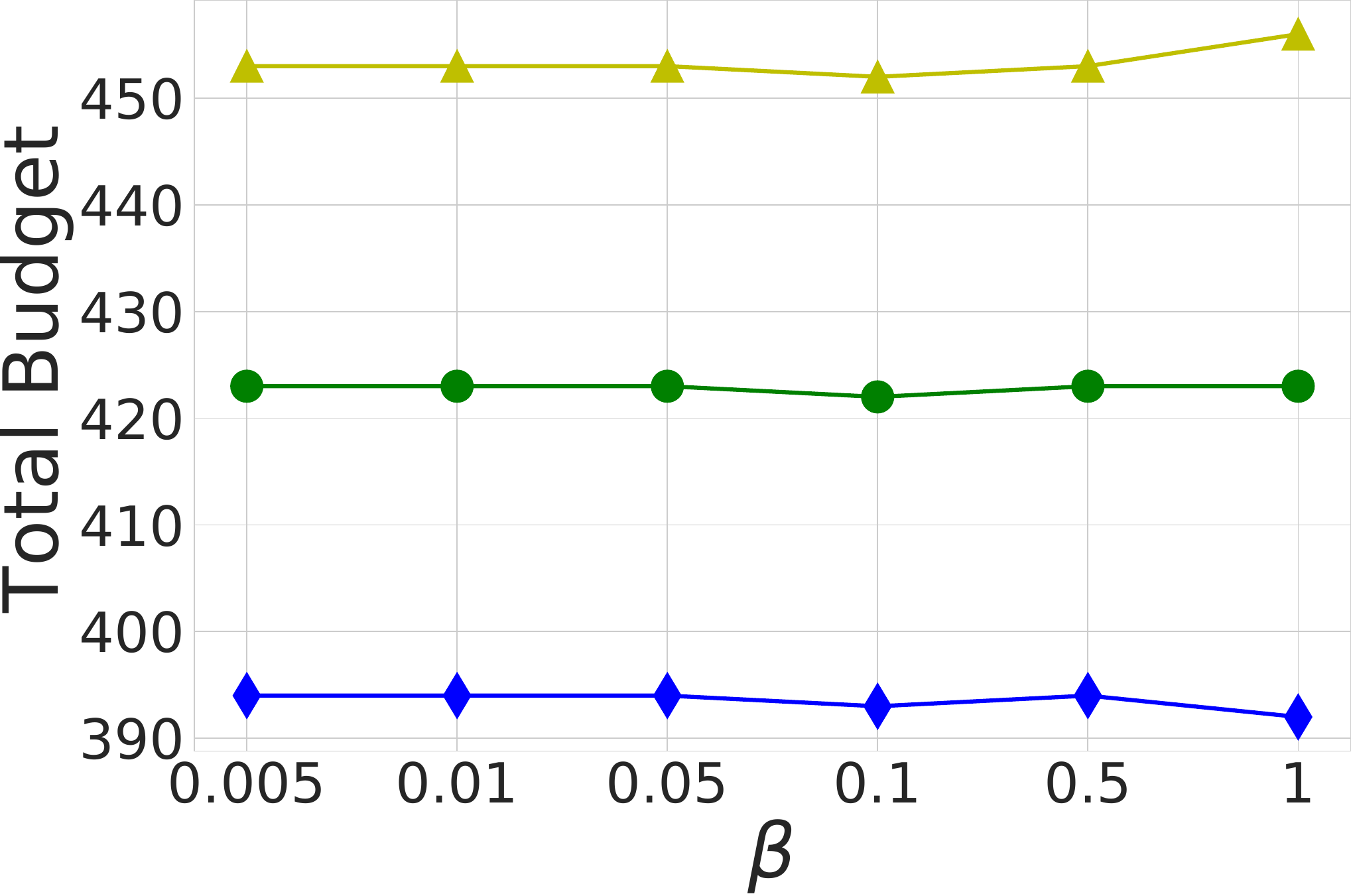}}
\endminipage\hfill
\endminipage\hfill
\vspace{-10pt}
\caption{Impact of $\beta$.}\label{fig:hyper_beta}
\end{figure}

\begin{figure}[!htbp]
\minipage{0.49\textwidth}
\centering
\minipage{0.49\textwidth}
\centering
\subfigure[Cora.] {    \includegraphics[width=\textwidth]{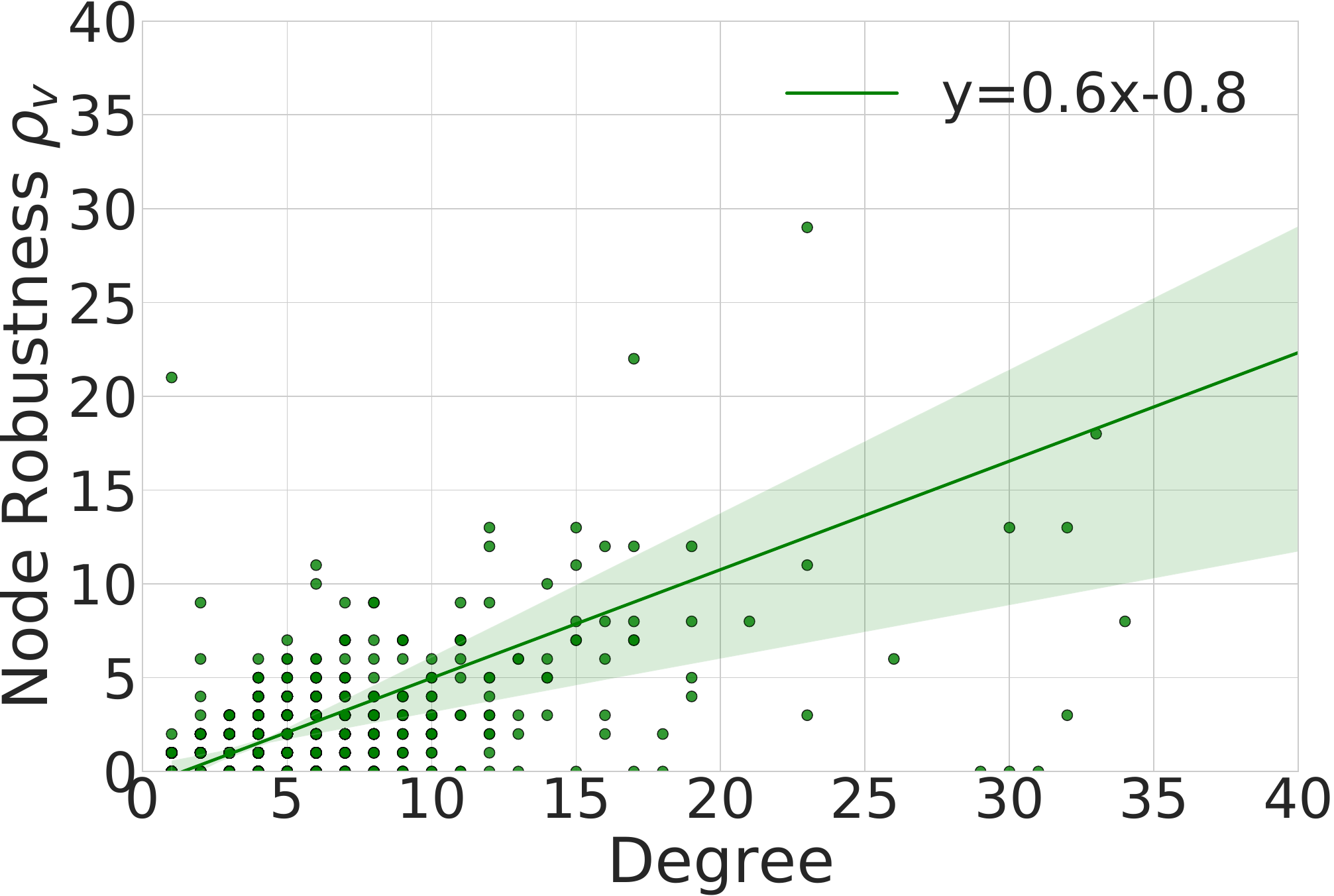}}
\endminipage\hfill
\minipage{0.49\textwidth}
\centering
\subfigure[Citeseer.] {    \includegraphics[width=\textwidth]{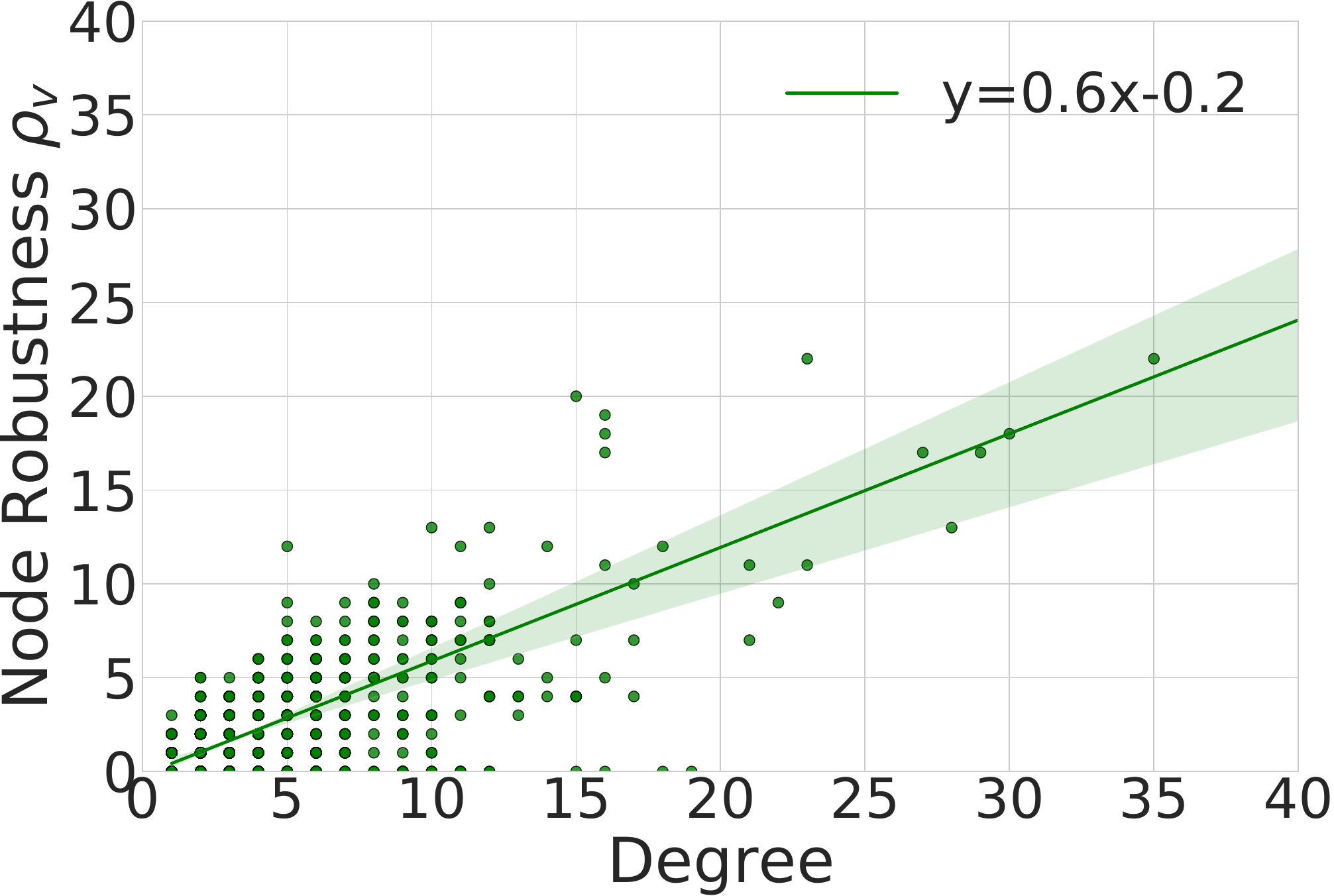}}
\endminipage\hfill
\endminipage\hfill
\vspace{-10pt}
\caption{Robustness versus degree.}\label{fig:rob2deg}
\end{figure}

\begin{figure}[!htbp]
\minipage{0.49\textwidth}
\centering
\minipage{0.49\textwidth}
\centering
\subfigure[Cora.] {    \includegraphics[width=\textwidth]{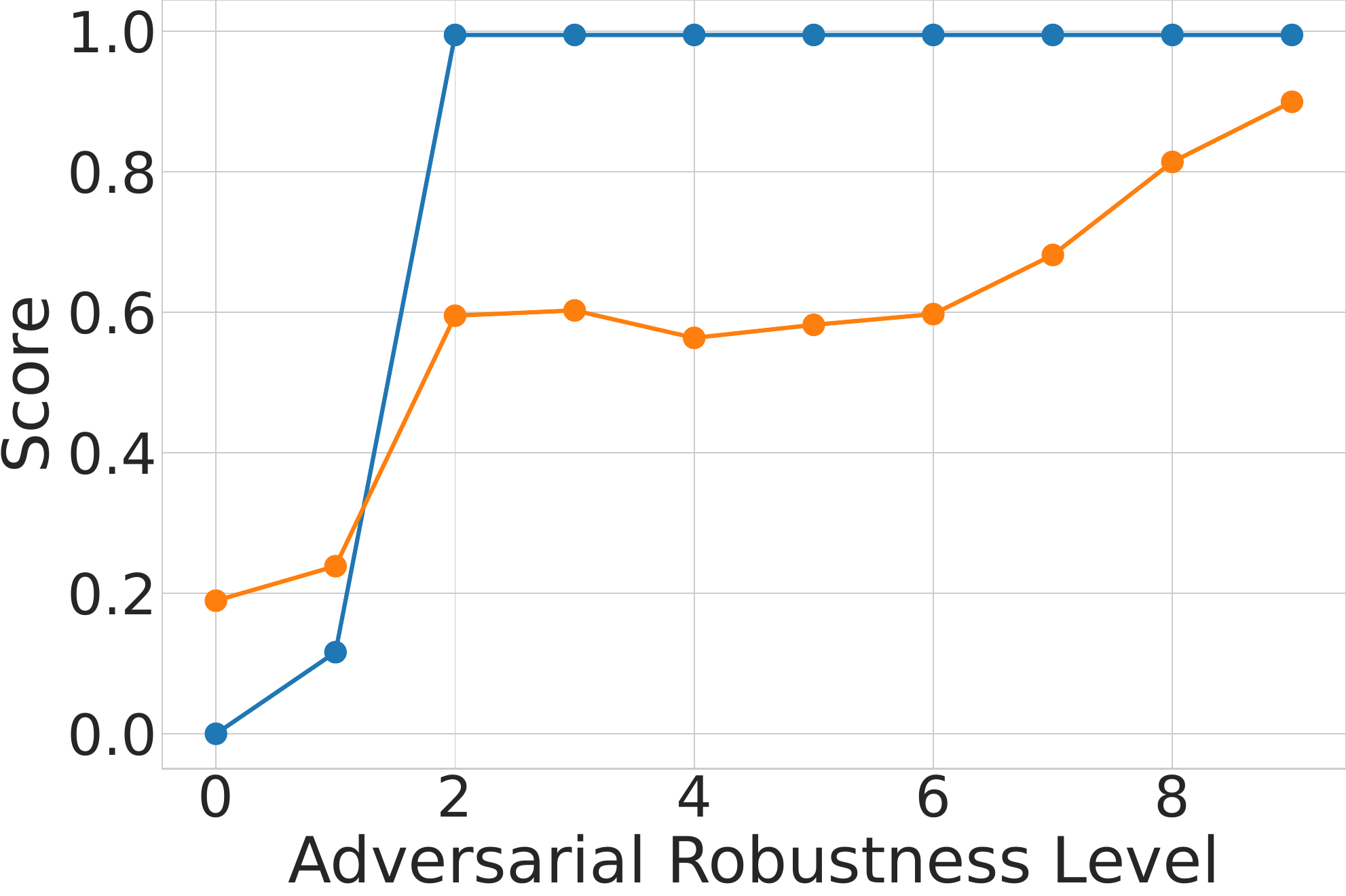}}
\endminipage\hfill
\minipage{0.49\textwidth}
\centering
\subfigure[Citeseer.] {    \includegraphics[width=\textwidth]{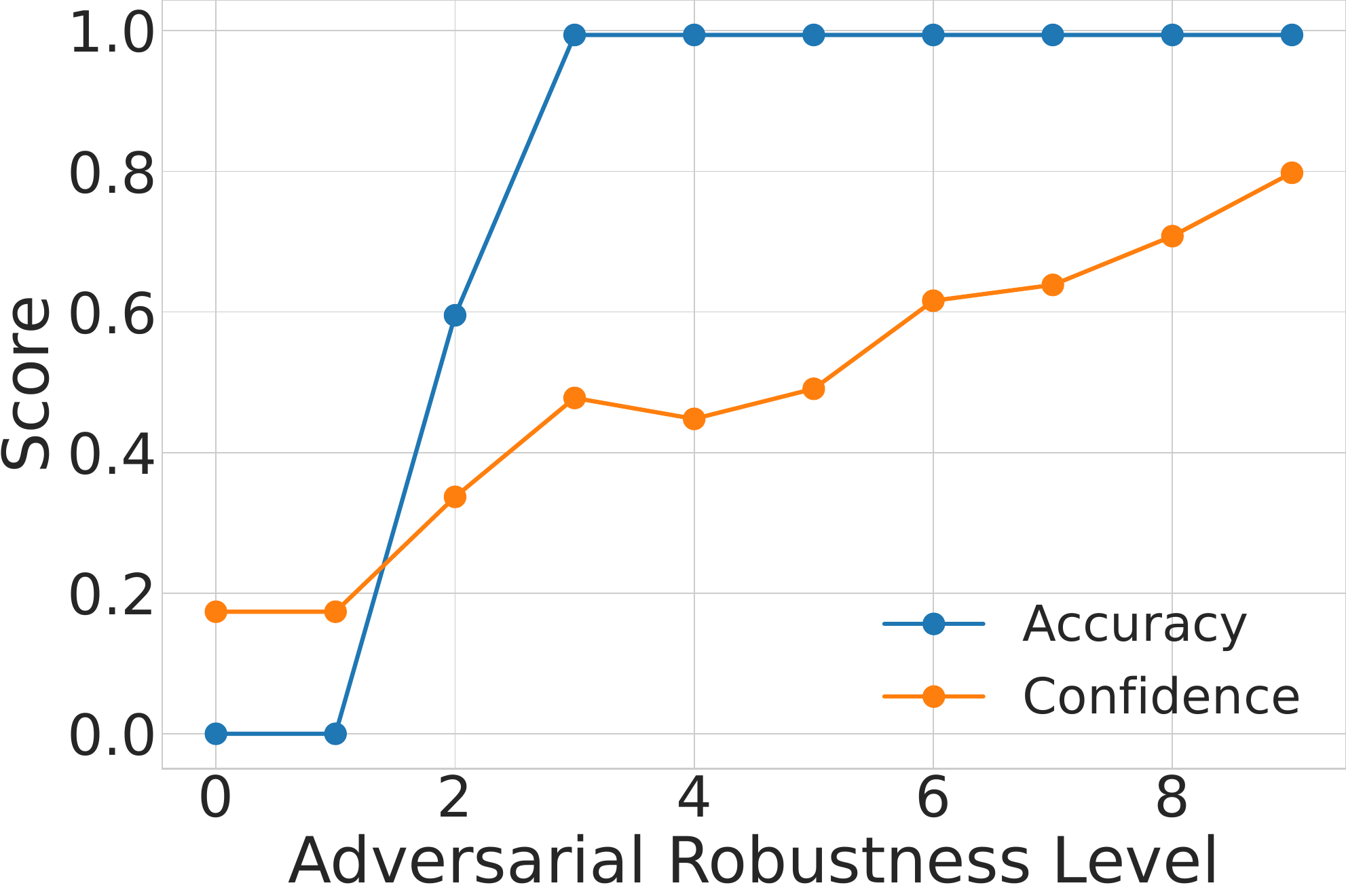}}
\endminipage\hfill
\vspace{-10pt}
\caption{Robustness versus uncertainty.}\label{fig:rob2conf}
\endminipage\hfill
\end{figure}

\begin{table}[]
\caption{The total budget of our attack models with/without the proposed dynamic PGD and Initialization.}
\vspace{-5pt}
\begin{tabular}{|c|ccc|}
\hline
Dataset  & W/o DPGD                  & W/o Init & MiBTack       \\ \hline
Cora     & 385                       & 364      & \textbf{330}  \\ \hline
Citeseer & 484                       & 459      & \textbf{444}  \\ \hline
Polblogs & 2038                      & 1967     & \textbf{1961} \\ \hline
Pubmed   & 357 & 353      & \textbf{348}  \\ \hline
\end{tabular}
\label{tab:ablation}
\vspace{-10pt}
\end{table}

\subsection{Model Analysis}

\textbf{Ablation Study}. We conduct an ablation study to evaluate the necessity of the components of our attack model. Take GCN as example, we report the total budget of our dynamic projected gradient descent and exploration operation as shown in Table~\ref{tab:ablation} (ACC are all 0).  
Specifically, we compare our full model \textbf{MiBTack} with the variant (\textbf{W/o DPGD}) using greedy method to replace the dynamic PGD and the model without initialization (\textbf{W/o Init}). 
One can observe that full model MiBTack significantly behaves better than W/o DPGD, suggesting that  our dynamic PGD can improve the combinatorial optimization problem and the challenge of non-convex constraint optimization under PGD can be alleviated by our designs. Meanwhile, compared to W/o Init, it can be seen that our customized initialization also benefits the proposed MiBTack.

\textbf{Impact of $P$ (patience).} Fig.~\ref{fig:hyper_P} demonstrates how the proposed attack methods perform against three GNNs when patience $P$ increases under three datasets. One can observe that the proposed model obtains lower minimum budget with the increasing of patience $P$, and only requires a few of the trials to converge.  
This is because $P$ is the number of maximal iteration after crossing the decision boundary, and a larger $P$ will lead to a better adjusting of $\Delta_v$ iteratively. Meanwhile, our model can quickly converge below 200 iterations, indicating the efficiency of our model.

\textbf{Impact of $\alpha$}. Hyperparameter $\alpha$ is the step size for updating $\boldsymbol{{\delta}}_v$ in Eq.~\ref{eq:PGD_gd}. Take the datasets Cora and Citeseer as examples, the results are reported in Fig.~\ref{fig:hyper_alpha}. As seen, there exists an optimal $\alpha$ that delivers the minimum budget. This is because our methods with too large step size $\alpha$ may fail to converge to a good solution. When step size $\alpha$ is too small, the elements of flipping operation vector have small values in continuous space and can only change the adjacency vector $\boldsymbol{a}_v$ slightly, which often leads to sub-optimal results due to the discrete structure of graph data.

\textbf{Impact of $\beta$}. We test the impact of hyperparameter $\beta$, which is the step size for updating budget $\boldsymbol{{\Delta}}_v$. As shown in Fig.~\ref{fig:hyper_beta}, we can observe that, basically, our framework is stable when $\beta$ is within the range from $\{0.005, 0.01, 0.05, 0.1, 0.5, 1\}$. This is because that the step size of $\boldsymbol{{\Delta}}_v$ is expected to only affect the time of convergence.

\vspace{-10pt}
\subsection{Node Robustness Analysis}\label{sec:5.5}
For each node $v$, we use the minimum budget $\Delta^{\star}_v$ generated by our MiBTack as node robustness $\rho_v$.

% \textbf{Visualization.}
% To provide a more intuitive evaluation, we conduct visualization experiment for target nodes on Cora and Citeseer dataset, showing their uncontaminated node embedding and node robustness $\rho_v$. We plot the learned embeddings of GCN using t-SNE, and the results are shown in Fig.~\ref{fig:visual}, in which different colors mean different labels, and the size of node indicates the calculated $\rho_v$. We can see that the size of node, namely node robustness, becomes larger with the increasing of its distance to the boundaries between different classes. 
% This hints that our node robustness can reflect the distance of target node to decision boundaries.
% % Thus a more precise minimum budget can better approximate the distance of target node to decision boundaries.

\textbf{Relationships of node robustness and degree.}
We analyze how the structure of the target node, i.e., its degree, affects the robustness. For each target node $v$ in Cora and Citeseer, we plot its degree and the node robustness $\rho_v$ found by MiBTack, then plot the corresponding linear regression line and equation as shown in Fig.~\ref{fig:rob2deg}. 
There are some observations: First, there is a positive correlation between degree and $\rho_v$ as expected in existing works~\cite{nettack}. The high degree nodes are harder to be perturbed to cross decision boundaries. Second, there exists a mass of nodes under regression line, they only need the perturbation edges with fewer than half of neighbors to mislead GCN, showing high vulnerability. 
% The reason can be attributed to the sum/mean based aggregation functions, which are proved to be distorted arbitrarily by several extreme outliers \cite{Chen2021UnderstandingSV,Geisler2020ReliableGN}. 

\textbf{Relationships of node robustness and uncertainty.}\label{sec:4.rob2prob}
The uncertainty of GNNs’ predictions, indicating how much we should trust our GNN, is crucial for their deployment in many risk-sensitive applications. Existing works often use the predicted probability (i.e, confidence) as the uncertainty indicator. Here we study the relationship between adversarial robustness and confidence together with accuracy. Following~\cite{Qin2020ImprovingCT}, we rank the input node according to their node robustness ($\rho_v$) and then divide the dataset into 10 equally-sized subsets. For each adversarial robustness subset, we compute accuracy and the average confidence score of the predicted class as shown in Fig.~\ref{fig:rob2conf}. One can clearly see that both accuracy and confidence increase with the adversarial robustness of the input data. With the increase of adversarial robustness level, compared to accuracy, the confidence becomes higher first then consistently lower. This indicates that GCN tends to give over-confident predictions for the easily attacked nodes, but give under-confident predictions otherwise, which fits nicely with the conclusion of existing work (GNNs are under-confident)~\cite{Wang2021BeCT}. Moreover, it makes a new interesting observation: There may exist a turning point for GNNs from over-confident to under-confident. This observation may help improve the confidence calibration for GNNs.

% \vspace{-pt}
\section{Conclusions}
\vspace{-5pt}
  In this paper, we introduce the first study on the minimum-budget topology attack on GNNs, to find the smallest perturbation for successful attack.
  We propose an effective attack model MiBTack based on the  differentiable dynamic projected gradient descent (PGD). Our MiBTack can effectively find the minimum perturbations for cross above decision boundary by differentiable dynamic PGD, solving the inherent intractable non-convex constrained optimization. The experimental results show that MiBTack can achieve 100\% attack success rate with minimum perturbation edges. Besides, we use these obtained minimum budget to study the relationships between robustness, topology and uncertainty.
  An interesting direction for future work is to extend our MiBTack to black-box setting.

\begin{acks}
This work is supported in part by the National Natural Science Foundation of China (No. U20B2045, 62192784, 62172052, 62002029, U1936104).
\end{acks}

\bibliographystyle{ACM-Reference-Format}
\bibliography{WWW2023}

\clearpage
\appendix

\section{Initialization}\label{app:initial}
The attack performance of our MiBTack may rely on the initial status $\boldsymbol{a'}^{(0)}_v = \boldsymbol{a}_v + \boldsymbol{\delta}^{(0)}_v$. While in multi-classification, the initial attack direction may not be optimal, namely dose not point to the closest decision boundary.
Specifically, in the first iteration, the initial adjacency $\boldsymbol{a'}^{(0)}_v=\boldsymbol{a}_v$ often has small and similar prediction confidence on all wrong classes $c(c \neq y_v)$, e.g., node 352 in Cora dataset has $f_{\theta^{*}}(\boldsymbol{a'}^{(0)}_v)$ [0.0133, 0.0039, 0.0215, 0.0361, 0.0130, 0.0550] for $c \neq y_v$ and 0.8571 for $y_v$. The $\mathcal{L}$ will update the perturbation vector along the direction to the wrong class with the largest prediction confidence $\max _{c \neq y_{v}} f_{\theta^{*}}(\boldsymbol{a'}^{(0)}_v)$. Obviously, the wrong class with largest confidence (0.0550) may only have little superiority than other wrong classes so it probably fails to indicate the closest decision boundary in multi-class classification. Note that the superiority of the wrong class picked in first iteration can be inherited under $\mathcal{L}$, guiding target node to cross a non-closest decision boundary with more perturbation edges wasted. 

However, it is time-consuming to precisely search the closest decision boundary, since the attacker needs to compute the distance (i.e., minimum budgets) of starting point to each decision boundary $\mathcal{B}_{c}$ for class $c \in \{1, \cdots, C\} \setminus \{y_v\}$. 
Here we estimate the closest decision boundary in multi-classification by a fast one-step attack instead of the time-consuming brute force, to trade off performance and efficiency. Then we initialize the staring point by slightly moving to the specified closest decision boundary, aiming to guide the generated attack trajectory to cross this boundary.

Clearly, this reminds us that precisely estimating the closest decision boundary beforehand and specifying it to guide the later attack generation can help generate a more unnoticeable attack.
However, given target node $v$, the precise way to search the closest decision boundary is to perform minimum-budget attacks towards each decision boundary $\mathcal{B}_c$ ($c \in \{1, \cdots, C\} \setminus \{y_v\}$), and choose the boundary with minimal budget as the closest one $\mathcal{B}_{c^\star}$, which is time-consuming. To trade off performance and efficiency, we simplify the minimum-budget attack for each class $c$ by limiting only flipping one edge, namely one-step attack, yielding a one-step perturbation $\boldsymbol{\delta}^{c}_v$ towards $c$ as:
\begin{align}\label{eq:exp_1}
      \boldsymbol{\delta}^{c}_v = \mathop{\rm arg min}\limits_{\boldsymbol{\delta}_v}  \quad & \mathcal{L}_{CW}(\boldsymbol{a}_v+\mathcal{T}(\boldsymbol{\delta}_v),c),\\
      = \mathop{\rm arg min}\limits_{\boldsymbol{\delta}_v} & \quad
      f_{\boldsymbol{\theta}^*}(\boldsymbol{a}_v+\mathcal{T}(\boldsymbol{\delta}_v))_{y_v} -  f_{\boldsymbol{\theta}^*}(\boldsymbol{a}_v+\mathcal{T}(\boldsymbol{\delta}_v))_{c}, \notag\\
    s.t. \quad & \|\boldsymbol{\delta_v}\|_0 \leq 1 \notag.
\end{align}
where $\boldsymbol{\delta}^{c}_v \in [\delta^{c}_{v1}, \delta^{c}_{v2}, \cdots, \delta^{c}_{vN}] \in [0,1]^N$ can be  found by minimizing the difference of confidence between $y_v$ and $c$. 
Then we approximately choose the most potential class $c^\star$ by comparing the decrease loss under each $c$:
\begin{equation}\label{eq:exp_2}
    c^\star = \mathop{\rm arg\ max}\limits_{c}  \mathcal{L}_{CW}(\boldsymbol{a}_v + \boldsymbol{\delta}^{c}_v, c).   
\end{equation}
Then we initialize the perturbation $\boldsymbol{\delta}^{(0)}_v$ as $\boldsymbol{\delta}^{c^\star}_v$, leading to a better starting point $\boldsymbol{a'}_v^{(0)}=\boldsymbol{a}_v+ \mathcal{T}(\boldsymbol{\delta}^{(0)}_v)$ which tends to have a significantly higher confidence on $c^\star$. 
% As shown in Fig.~\ref{fig:explore_toy}, the blue starting point with red outline has slightly higher confidence on green $c_3$ than purple $c_2$ ($f_{\boldsymbol{\theta}^*}(\boldsymbol{a'}^{(0)}_v)_{c_3}=0.1$), the MiBTack with $\mathcal{L}_{CW}$ tends to inject a neighbor with class $c_3$, leading to a shortest monotonous green attack trajectory to the closest boundary $\mathcal{B}_{c_3}$.  

\section{Algorithm}\label{app:alg}

Here we provide the pseudo-code in Algorithm \ref{alg} for the whole process of MiBTack. 
Given a target node $v$, we initialize the starting adjacency vector $\boldsymbol{\alpha}^{\prime(0)}_v$ by one-step attack for each class $c \in \{1,\cdots,C\}$ in Line 2-3. The training procedure of the perturbation vector $\boldsymbol{{\delta}}_v$ is presented in Line 4-11, which learns $\boldsymbol{{\delta}}_v$ with PGD (Line 5-7), then dynamically adjusts $\Delta_v$ and other parameters (Line 8-11). Lastly, we output the attacks when $P \leq 0$ or $\| \boldsymbol{\delta}_v^{\star} \|_0 \leq 1$. Intuitively, before attack, we first estimate the closest decision boundary, and then use it to initialize the starting point to guide the attack trajectory to cross the closest decision boundary. Then we search for the most adversarial topology attacks under the current budget with PGD and then the budget is enlarged or reduced based on whether these attacks succeed.  Through repeatedly crossing the green decision boundary, MiBTack can find the optimal budget $\Delta^\star_v$, namely the green dotted circle centered at node $v$, which is tangent to the green decision boundary.

\section{Space  Consumption Analysis}\label{app:space}
With the burden of a dense adjacency matrix, the gradient-based topology attacks, including our MiBTack, usually have the space complexity $O\left(N^2\right)$~\cite{geisler2021robustness}, where $N$ is the number of nodes. Fortunately, there are several works overcome this limitation~\cite{Feng2020ScalableAA,geisler2021robustness}. For example, PRBCD~\cite{geisler2021robustness} based on  Randomized Block Coordinate Descent (R-BCD) only has  linear complexity w.r.t. the budget $\Delta$. 
It worth pointing out that the space complexity of our MiBTack can be further reduced by combining with these methods.
We leave extending our MiBTack with Randomized Block Coordinate Descent as our future work.

\begin{algorithm}[!t]
\caption{MiBTack}
\label{alg}
\begin{algorithmic}[1]
\REQUIRE 
The target node $v$, 
the trained GNN model $f$, 
patience $P$, 
the initial step size $\alpha$ and $\beta$. \\
\ENSURE The minimal-budget adversarial perturbations $\boldsymbol{\delta}^{\star}_v$.
\STATE $i \leftarrow 1$, $\Delta^{(0)}_v=1$.\\
\STATE Approximate the closet boundary $c^\star$ via Eq.~\ref{eq:exp_2}, then initialize $\boldsymbol{\delta}_v^{(0)}$ with $\boldsymbol{\delta}^{c^\star}_v$.\\
\WHILE{$P \leq 0$}
\STATE Update $\boldsymbol{\delta}_v^{(i-1)}$ to $\Tilde{\boldsymbol{\delta}}_v^{(i)}$ with gradient descent with Eq.~\ref{eq:PGD_grad} and \ref{eq:PGD_gd}.\\
\STATE Project $\Tilde{\boldsymbol{\delta}}_v^{(i)}$ for the constraint of  $\|\boldsymbol{\delta}_v^{(i)}\|_0 \leq \Delta^{(i)}_v$ by Eq.~\ref{eq:PGD_proj}.\\
\STATE Obtain the perturbed adjacency vector by $\boldsymbol{a}^{\prime(i)}_v \leftarrow \boldsymbol{a}_v + \mathcal{T}(\boldsymbol{\delta}_v^{(i)})$.
\IF{$\mathcal{L}({\rm Proj}_{\{0,1\}^N}[\boldsymbol{a'}^{(i)}_v])<0$}
\STATE Reduce budget to $\Delta^{(i+1)}_v$ with Eq.~\ref{eq:dyn_shrink}.
\IF{$\|\boldsymbol{\delta}_v^{{(i)}}\|_0<\|\boldsymbol{\delta}_v^{\star}\|_0$}
\STATE $\boldsymbol{\delta}_v^{\star} \leftarrow \boldsymbol{\delta}_v^{{(i)}}$.
\ENDIF
\ELSE
\STATE Enlarge budget to $\Delta^{(i+1)}_v$ with Eq.~\ref{eq:dyn_enlarge}.
\ENDIF
\STATE Reduce patience $P$ and step size $\alpha$ and $\beta$ if the perturbed node has reached decision boundary.
\ENDWHILE 
\RETURN Minimum perturbation $\boldsymbol{\delta}^{\star}_v$  and $\Delta^{\star}_v=\|\boldsymbol{\delta}_v^{\star}\|_0$.
\end{algorithmic}
\end{algorithm}

\section{Future Details On Experiment}\label{app:detail_exp}
\subsection{Dataset}\label{app:data}

The dataset characteristics are shown in Table~\ref{tab:dataset}, and we only consider the largest connected component as in~\cite{nettack}. We split the network into labeled (20\%) and unlabeled nodes (80\%). We further equally split the labeled nodes into training and validation sets to train our surrogate model. The datasets used in this paper can be found in \url{https://github.com/DSE-MSU/DeepRobust}.

\begin{table}
\caption{Dataset statistics.}
\begin{tabular}{|c|r|r|r|r|}
\hline
              & Polblogs  & Cora       & Citeseer     & Pubmed       \\ \hline
\#Nodes        & 1,222  & 2,485       & 2,110         & 19,717          \\ \hline
\#Edges        & 16,724   & 5,069       & 3,668         & 44,325         \\ \hline
\#Features        & -   & 1,433       & 3,703         & 500                 \\ \hline
\#Classes         & 2  & 7          & 6            & 3                      \\ \hline
% Tr/Val/Te & 247/249/1988  & 210/211/1688 & 1971/1972/15774 & 121/123/978  \\ \hline
\end{tabular}\label{tab:dataset}
\end{table}

% \subsection{Implementations of Target Models} \label{app:gnns}
% To validate the generalization ability of our proposed attacker, we choose three popular graph neural networks: 
% (1) GCN~\cite{GCN} is a representative GNN and learns on graph structures using convolution operations. We train a 2-layer GCN with learning rate 0.01, where the number of units in hidden layer is 16. In addition, the dropout rate is 0.5, weight decay is $5e-4$. 
% (2) SGC~\cite{Wu2019SimplifyingGC} simplifies GCNs through successively removing nonlinearities and collapsing weight matrices between consecutive layers.  For SGC, the learning rate is 0.01, the number of units is 16,  the dropout rate is 0.5, and weight decay is $5e-6$. 
% (3) APPNP~\cite{Klicpera2019PredictTP} further improves GCN by leveraging residual connection to preserve the information of raw features. The learning rate of APPNP is 0.01, the number of units in hidden layer is 64, the dropout rate is 0.5, weight decay is $5e-6$. 

\subsection{Implementations of Attack Models} \label{app:codes}
All baselines are initialized with same parameters suggested by their papers and we also further carefully turn parameters to get optimal performance. For the greedy-based baselines (i.e., Rand, DICE, DICE-t, FGA and Nettack), we set the maximum epoch as 1000, where they can consistently add perturbations until target node $v$ is misclassified or the number of perturbed edges is more than 1000.  For our model, we set the patience $P$ as 800, where the step size $\alpha$ for updating $\boldsymbol{\delta}^{(i-1)}_v$ is 1.0 and $\beta$ for updating $\Delta^{(i)}_v$ is 0.1. 
% For the reproducibility, we also provide the codes of used GNN models and baselines.

% \subsection{Implementations of Variant Models} \label{variant}
% For variants MiBTack-J and MiBTack-R, we replace the default surrogate GCN as Jaccard-GCN and RGCN.

\subsection{Experiments Settings} \label{app:exp_setting}
All experiments are conducted with following setting:
\begin{itemize}
    \item Operating system: CentOS Linux release 7.7.1908(Core)
    \item CPU: Intel(R) Xeon(R) Silver 4210 CPU @ 2.20GHz
    \item GPU: GeForce RTX 2080 Ti
    \item Software versions: Python 3.8.5; Pytorch 1.8.0; Numpy 1.19.2; SciPy 1.5.4; NetworkX 2.5; Scikit-learn 0.23.2; dgl 0.5.3; torch-sparse 0.6.12;
\end{itemize}

% \subsection{Visualization}
% To provide a more intuitive evaluation, we conduct visualization experiment for target nodes on Cora and Citeseer dataset, showing their uncontaminated node embedding and node robustness $\rho_v$. We plot the learned embeddings of GCN using t-SNE, and the results are shown in Fig.~\ref{fig:visual}, in which different colors mean different labels, and the size of node indicates the calculated $\rho_v$. We can see that the size of node, namely node robustness, becomes larger with the increasing of its distance to the boundaries between different classes. 
% This hints that our node robustness can reflect the distance of target node to decision boundaries.
% % Thus a more precise minimum budget can better approximate the distance of target node to decision boundaries.

% \begin{figure}
%     \centering
%     \includegraphics[width=\textwidth]{fig/visual_cora.pdf}
%     \caption{Caption}
%     \label{fig:visual}
% \end{figure}

% \begin{figure}
% \minipage{0.49\textwidth}
% \centering
% \subfigure[Cora.] {    \includegraphics[width=\textwidth]{fig/visual_cora.pdf}}
% \endminipage\hfill
% \minipage{0.49\textwidth}
% \centering
% \subfigure[Citeseer.] {    \includegraphics[width=\textwidth]{fig/visual_citeseer.pdf}}
% \endminipage\hfill
% \vspace{-10pt}
% \caption{Visualization of node robustness.}\label{fig:visual}
% \end{figure}

\end{document}